\documentclass[11pt]{article}

\usepackage[]{acl}

\usepackage{times}
\usepackage{latexsym}

\usepackage[T1]{fontenc}

\usepackage[utf8]{inputenc}

\usepackage{microtype}

\usepackage{inconsolata}

\usepackage{graphicx}

\usepackage[textsize=footnotesize]{todonotes}

\newcommand{\boxed}[1]{\todo[inline,color=olive!20!white]{#1}} 

\usepackage{graphicx}
\usepackage[table]{xcolor}
\usepackage{booktabs} 
\usepackage{cleveref}

\usepackage{booktabs}
\usepackage{xspace}
\usepackage{makecell}

\usepackage{amssymb}   
\usepackage{stmaryrd}  
\usepackage{wasysym}   

\usepackage{fvextra}

\newcommand{\commandAplus}{Command A+\xspace}

\newcommand{\nst}{North Small Translate\xspace}

\title{North Small Translate: Advanced Cost-Effective Translation (Cohere CAT+)}

\author{
  Tom Kocmi\textsuperscript{*}
  \And
  Alexandre Bérard
  \And
  Phil Blunsom
  \And
  Samuel Cahyawijaya
  \AND
  Shaun Cassini
  \And
  Nicholas Frosst
  \And
  Ona de Gibert
  \And
  Aidan Gomez
  \And
  Nithya Govindarajan
  \AND
  Shun Kiyono
  \And
  Olivia Lasche
  \And
  Lawrence Rogers
  \And
  Kelly Marchisio
  \And
  Nikita Moghe
  \AND
  Yash More
  \And
  Camila Moran-Hidalgo
  \And
  Yiyang Nan
  \And
  Michael Sachs
  \And
  Trisha Starostina
  \AND
  Daan van Stigt
  \And
  Spencer Rarrick
  \And
  Sebastian Vincent
  \And
  Ivan Zhang
  \AND
  Cohere \\
  \textsuperscript{*}\href{mailto:kocmi@cohere.com}{kocmi@cohere.com}
}

\begin{document}
\maketitle

\begin{abstract}

We present \textit{\nst}, an open-weight, LLM-based machine translation (MT) model with instruction-following capabilities built on the same foundation as Cohere's \textit{\commandAplus}, a mixture-of-experts architecture with 25 billion active parameters out of 218 billion total parameters. \textit{\nst} is trained using difficulty sampling to obtain challenging documents and a five-step training protocol combining supervised fine-tuning, direct preference optimization, and online reinforcement learning.
We prioritized throughput through a non-reasoning base model and supplemented with optional agentic capabilities to unlock translation quality gains.
\textit{\nst} is trained to perform MT-related tasks, including post-editing and quality estimation, as well as related tasks such as general instruction following.  
The model achieves top MT performance across 50 languages in the class of models under 1T parameters, with no need to run expensive reasoning at inference time.

\end{abstract}

\begin{table*}[t]
\centering
\resizebox{\textwidth}{!}{%
\begin{tabular}{llrrrrr}
\toprule
 & \makecell{Params\\total (active)} & \makecell{GEMBA\\WMT26} & \makecell{xCOMETxl\\WMT24++} & \makecell{Terminology} & \makecell{Long\\context} & \makecell{Structured\\translation} \\
\midrule
\cellcolor[HTML]{D9F0D3}North Small Translate (Agentic) & 218B (25B) & {\cellcolor[HTML]{668F76}} 84.4 & {\cellcolor[HTML]{9ECFA2}} 80.7 & {\cellcolor[HTML]{72A683}} 87.9 & {\cellcolor[HTML]{8ABF94}} 47.8 & {\cellcolor[HTML]{6C9B7D}} 93.3 \\
\cellcolor[HTML]{D9F0D3}North Small Translate & 218B (25B) & {\cellcolor[HTML]{6B997B}} 83.6 & {\cellcolor[HTML]{A5D4A8}} 80.6 & {\cellcolor[HTML]{6D9C7D}} 89.7 & {\cellcolor[HTML]{83B98F}} 48.9 & {\cellcolor[HTML]{6A977B}} 93.7 \\
Mistral Large 3 & 675B (41B) & {\cellcolor[HTML]{7EB58C}} 81.6 & {\cellcolor[HTML]{E2D2E7}} 78.3 & {\cellcolor[HTML]{699479}} 91.2 & {\cellcolor[HTML]{E6D8EA}} 20.0 & {\cellcolor[HTML]{6A977B}} 93.8 \\
Qwen 3.5 397B & 397B (17B) & {\cellcolor[HTML]{80B68D}} 81.6 & {\cellcolor[HTML]{EBDDED}} 78.5 & {\cellcolor[HTML]{75AC86}} 87.0 & {\cellcolor[HTML]{E4D6E9}} 19.5 & {\cellcolor[HTML]{C9B2D3}} 59.4 \\
DeepL NextGen &  & {\cellcolor[HTML]{83B98F}} 81.4 & {\cellcolor[HTML]{668F76}} 81.8 & {\cellcolor[HTML]{BD9FC7}} 47.8 & {\cellcolor[HTML]{668F76}} 56.5 & {\cellcolor[HTML]{E3D4E8}} 64.5 \\
Gemma 4 31b & 31B (31B) & {\cellcolor[HTML]{A3D3A6}} 79.5 & {\cellcolor[HTML]{FAFAFA}} 79.2 & {\cellcolor[HTML]{668F76}} 92.1 & {\cellcolor[HTML]{E4D6E9}} 19.4 & {\cellcolor[HTML]{668F76}} 95.1 \\
Inkling Small & 276B (12B) & {\cellcolor[HTML]{B7DEB6}} 78.5 & {\cellcolor[HTML]{F2E8F3}} 78.7 & {\cellcolor[HTML]{699479}} 91.1 & {\cellcolor[HTML]{ECE0EE}} 21.5 & {\cellcolor[HTML]{73A884}} 91.3 \\
Muse Glimmer 30B & 30B (30B) & {\cellcolor[HTML]{BEE3BC}} 78.2 & {\cellcolor[HTML]{C4ACCE}} 77.7 & {\cellcolor[HTML]{6C9A7C}} 90.1 & {\cellcolor[HTML]{ECF7EA}} 32.5 & {\cellcolor[HTML]{72A683}} 91.7 \\
\cellcolor[HTML]{D9F0D3}Command A+ & 218B (25B) & {\cellcolor[HTML]{D9F0D5}} 76.5 & {\cellcolor[HTML]{84BA90}} 81.0 & {\cellcolor[HTML]{EEF7EB}} 68.7 & {\cellcolor[HTML]{F3F8F1}} 30.5 & {\cellcolor[HTML]{F3F9F2}} 73.7 \\
GLM 5.2 & 744B (40B) & {\cellcolor[HTML]{D9F0D5}} 76.5 & {\cellcolor[HTML]{F3EAF3}} 78.8 & {\cellcolor[HTML]{B5DDB5}} 78.2 & {\cellcolor[HTML]{F9FAF9}} 28.6 & {\cellcolor[HTML]{70A281}} 92.1 \\
Mistral Medium 3.5 & 128B (128B) & {\cellcolor[HTML]{E5F5E1}} 75.7 & {\cellcolor[HTML]{956D9C}} 76.7 & {\cellcolor[HTML]{6B997B}} 90.3 & {\cellcolor[HTML]{E4D5E8}} 19.3 & {\cellcolor[HTML]{699479}} 94.2 \\
GPT-OSS 120B & 117B (5B) & {\cellcolor[HTML]{F6F2F6}} 72.3 & {\cellcolor[HTML]{8C6693}} 76.6 & {\cellcolor[HTML]{8ABF94}} 83.8 & {\cellcolor[HTML]{EADCEC}} 20.8 & {\cellcolor[HTML]{6C9B7D}} 93.2 \\
\cellcolor[HTML]{D9F0D3}Command A Translate (2025) & 111B (111B) & {\cellcolor[HTML]{E7D9EB}} 70.0 & {\cellcolor[HTML]{9D73A4}} 76.9 & {\cellcolor[HTML]{E5D7EA}} 56.9 & {\cellcolor[HTML]{8C6693}} 0.0 & {\cellcolor[HTML]{D5EED1}} 79.5 \\
Google Translate &  & {\cellcolor[HTML]{D5C2DE}} 68.2 & {\cellcolor[HTML]{95C89B}} 80.8 & {\cellcolor[HTML]{8C6693}} 38.1 & {\cellcolor[HTML]{EBDFEE}} 21.3 & {\cellcolor[HTML]{8C6693}} 48.8 \\
Nemotron 3 Ultra 550B & 550B (55B) & {\cellcolor[HTML]{C9B2D3}} 67.1 &  & {\cellcolor[HTML]{83B98F}} 84.8 & {\cellcolor[HTML]{E5D7EA}} 19.7 &  \\
Qwen 3.6 27B & 27B (27B) & {\cellcolor[HTML]{8C6693}} 61.9 &  & {\cellcolor[HTML]{6E9F7F}} 89.2 & {\cellcolor[HTML]{F8FAF7}} 28.9 &  \\
\bottomrule
\end{tabular}
}
\caption{Aggregated results of our model against other top systems with under 1T parameters.}
\label{tab:general_results}
\vspace{-1em}
\end{table*}

\section{Introduction}

The landscape of machine translation (MT) has undergone a profound transformation in recent years, having almost fully transitioned to large language models (LLMs) that treat cross-lingual generation as a complex instruction-following task. The paradigm shift has unlocked unprecedented capabilities, allowing models to leverage broad contextual understanding, perform robust translation, and adapt translations to various requests. 

We introduce \textit{\nst} (submitted to the WMT General MT shared task as \textit{Cohere CAT+}), our best-performing MT system built upon the same foundation as Cohere's \textit{\commandAplus}~\citep{cohere2026commandaplus}. It is a sparse mixture-of-experts (MoE) transformer with 25 billion active parameters out of a total of 218 billion parameters. 

\textit{\nst} is built for efficiency.  While most LLMs rely on test-time reasoning to achieve state-of-the-art performance, long reasoning traces introduce prohibitive latency and computational overhead for production-level translation. Model throughput is a primary target for practical MT deployment, so we explicitly trained \textit{\nst} as a non-reasoning model. For latency-insensitive applications, our optional \emph{Agentic Translation} framework bridges the potential quality gap caused by the absence of intermediate reasoning steps.

\textit{\nst} achieves top-tier translation quality through five distinct training steps: (i) Coarse SFT to establish general non-MT instruction-following capabilities and extend the base model's language coverage; (ii) Fine-grained SFT using highly curated synthetic datasets to improve core MT quality and adjacent tasks like error detection and post-editing; followed by three steps focusing primarily on human-aligned translation quality: (iii) DPO; (iv) Online RL using an LLM-as-a-judge reward; and (v) a minimal final DPO with reduced learning rate to resolve edge-case regressions introduced in RL.

We participate in the WMT 2026 General MT shared task \citep{kocmi-etal-2026-findings}, Terminology shared task \citep{charkiewicz-etal-2026-findings}, and Automated MT Evaluation shared tasks with our model under name \textit{Cohere CAT+}. We focused our submission on a clean performance without relying on any specialized search heuristics or inference tricks such as Minimum Bayes Risk decoding \citep{freitag-etal-2022-high}. Neither did we use the Agentic translation.\footnote{Disclosure of conflict: the main author is also an organizer of the WMT General MT shared task.} One domain for General MT contains videos as sources, we used \textit{Cohere Transcribe}\footnote{\url{https://cohere.com/blog/transcribe}} to obtain textual transcripts for English videos, while relying on the official provided transcripts for Chinese and Czech source data.

\section{Training Details}

\subsection{Model Architecture}

Our model is built on top of \commandAplus{} \citep{cohere2026commandaplus}, a decoder-only sparse Mixture-of-Experts transformer \citep{vaswani2017attention} with 218B total parameters and 25B active.\footnote{\url{https://huggingface.co/CohereLabs/command-a-plus-05-2026-bf16}}

The list of pre-training languages is: 

\boxed{Arabic, Bengali, Czech, Danish, Dutch, English, Filipino, Finnish, French, German, Greek, Hebrew, Hindi, Indonesian, Italian, Japanese, Korean, Malay, Norwegian, Persian, Polish, Portuguese, Punjabi, Romanian, Russian, Simplified Chinese, Spanish, Swedish, Tamil, Telugu, Thai, Traditional Chinese, Turkish, Ukrainian, Urdu, Vietnamese.}

The list of languages extended in post-training:

\boxed{Albanian, Bulgarian, Catalan, Croatian, Estonian, Hungarian, Icelandic, Irish, Latvian, Lithuanian, Maltese, Serbian, Slovak, Slovenian.}

\subsection{Training Approach}

We build on top of findings from our previous models and also our previous Command-A-Translate \citep{kocmi-etal-2025-command} which showcased the impact of Direct Preference Optimization (DPO) \citep{10.5555/3666122.3668460} training step on MT quality.

We use five distinct training steps, each focused on a different aspect:

\begin{enumerate}
    \item Initial coarse SFT: to cover general instruction-following capabilities, and extend language coverage from pretrained 35 languages to 50 languages. We fine-tune four different models, and merge them \citep{cohere2025command}.
    \item Fine-grained SFT: focusing on improving the MT quality using primarily synthetic datasets and also improving MT related tasks such as error detection, terminology or post-editing.
    \item DPO: primarily focused on improving quality.
    \item Online RL: online RL with GSPO~\citep{zheng2025gspo} and LLM-as-a-Judge as the reward signal to polish performance and increase adequacy.
    \item Light targeted DPO with reduced learning rate: for fixing remaining problems, especially a few introduced by the RL step.
\end{enumerate}

\Cref{tab:training_steps_progression} highlights how the MT performance improved by each individual training step. We see significant gains with each step except for the RL. While this step improved some scores, especially adequacy, it degraded the performance over few low-resource languages, which is due to weakness of the LLM judge on these languages.

To fix the aforementioned regression, we introduced a final DPO step reduced learning rate and using only 5 warm-up steps. For this step, we focused only on a few problematic languages and capabilities.

\begin{table}[h]
\centering
\begin{tabular}{lc}
\hline
\textbf{Step} & \textbf{MT Quality}\\
\hline
Coarse SFT & 71.3\% \\
Fine-grained SFT & 76.2\%  \\
DPO & 80.1\%  \\
Online RL & 80.0\%  \\
Light DPO & 81.8\% \\
\hline
\end{tabular}
\caption{Training progression on the WMT25 evaluation with GEMBA-ESA averaged over 14 languages.}
\label{tab:training_steps_progression}
\end{table}

\subsection{Data Preparation}

While the main goal is to build a strong MT model, we additionally focused on building a strong instruction-following model that also excels in MT-related tasks, such as post-editing, terminology following, quality estimation, or structured translation.

For every training set, we checked that every translation pair was processed through a suite of automated verifiers designed to eliminate noisy or malformed text. Specifically, we enforced strict structural consistency by verifying line and paragraph alignment, ensuring that the source and target segments contained the exact same number of lines and paragraphs. Additionally, we applied language identification and filtered out instances of unwanted code-switching. Our filtering pipeline also removed segments containing broken Unicode characters and relied on various other internal classifiers to detect and discard anomalies.

\subsubsection{Quality of Machine Translation}

A primary challenge in curating training corpora for top-performing MT systems is the sparsity of genuinely difficult source material, as we highlighted in \citet{kocmi-etal-2025-command}. After our initial coarse SFT stage, the model translates without major errors for more than 90\% of our training documents. Consequently, randomly sampling source texts would yield a weak learning signal and fail to push the model's quality boundaries. 

We use difficulty sampling to focus our synthetic generation exclusively on problematic instances from which the model can learn. Building upon our previous work \citep{kocmi-etal-2025-command}, we extend difficulty sampling by using the initial base model with minimal post-training itself as the baseline model and keep only documents where the model produces major errors, or its translation quality is lower than the human reference for parallel data. 
For each source document, we randomly pick a high-resource target language and translate it with an early stage model. The generated translation is then automatically evaluated for errors with a judge. For monolingual data, if the translation is largely error-free, the source segment is classified as ``easy'' and removed from the candidate pool. For parallel data, we check if the translation is better than the human reference to deem it ``easy''. By aggressively filtering out easily translated segments, we distill a highly concentrated corpus of difficult texts. This ensures that our downstream generation of synthetic data is allocated to the documents where the model requires the most improvement.

To achieve top MT performance across all supported languages, we construct our training corpora using two distinct data synthesis and refinement strategies, which we call \emph{Best-per-Language Forward Translation} and \emph{Post-Edit Driven Preference Distillation}.
Both approaches are designed to maximize the quality and diversity of the training signal without relying on human-annotated parallel data, which we use solely in the first stage of training.

Best-per-Language Forward Translation relies on a standard forward translation approach combined with a Best-of-N selection. We use a set of top-performing language-specific expert models, where for each language we select the top-performing expert to generate translations for given source documents. To ensure broad multilingual coverage and explicitly improve cross-lingual transfer, we randomly assign the target language during generation. While the majority of the source texts are in English, a significant part consists of non-English sources, yielding a substantial volume of direct non-English to non-English translation pairs. The resulting translation candidates are then evaluated by a judge and accepted only if their score is higher than 70 GEMBA-ESA. We describe details of the scoring in \Cref{ssec:llm_as_a_judge} as we use this approach throughout the work.

Our second strategy, \emph{Post-Edit Driven Preference Distillation}, is the most critical of our data preparation pipeline and is the primary driver for our DPO stage. The core objective is to generate near on-policy training data in an offline setting, creating highly effective preference pairs that directly target the model's own weaknesses. The orchestration pipeline operates in three distinct steps: 

\begin{enumerate}
    \item Generate a draft translation using the current candidate model.
    \item A separate judge model identifies and annotates error spans in the generated draft. If no error is present, the sample is deemed ``easy'' and dropped from the training data.
    \item A third model acts as an automated post-editor, taking the draft and fixing the errors.
    \item We iterate steps 2-3 until no major error is discovered, or we pass 10 iterations.
\end{enumerate}

This multi-model orchestration process yields a perfectly aligned preference pair: the original, flawed output from the candidate model serves as the worse completion, with the corrected, post-edited version as the preferred completion. \textit{Because the broken translation originates directly from the candidate model, the resulting dataset closely approximates on-policy MT data. This allows the offline DPO training to function with the precision of online reinforcement learning, penalizing the model's actual failure modes.} This orchestration approach was instrumental in fixing persistent, model-specific errors and driving the most significant quality improvements observed during our DPO stage.

\subsubsection{Instruction Following}
To ensure our MT model has acceptable general instruction following abilities which would strengthen the translation-related tasks and following specific MT instructions, we included non-MT related instruct data in the first SFT stage. We used some of the English-only single-turn datasets that were used to train \commandAplus{}. Additionally, we included multilingual datasets using general-purpose multilingual prompts. We have removed reasoning traces from all reused datasets. We furthermore used all the aforementioned data in the second SFT stage, but significantly downsampled.

\subsubsection{Structured Translation}

``Structured translation'' is the process of translating natural language embedded in data formats such as JSON or HTML without breaking their underlying syntax. It is a critical feature especially for translating webpages, documents or training data.

While general instruction following datasets described above contain various formats, we focus in structured translation solely on JSON and HTML formats.
We prepare data two ways: naturally occurring structured data and combining paragraphs or sentences from existing high quality MT datasets into diverse structures. For the former, we relied on our internal datasets containing JSONs and we then extracted the HTML data from various web based datasets. For the latter part, this was done using top segments from our high quality sentences corpora. After constructing this source data, we translated them using an internal model into 35 languages. These translations were done in a field-by-field basis as well as complete structure translations. 
When preparing the training data, we added prompt instruction controllability of levels of translation, from translating all structures in the source up to only focusing on translation of few keys in JSONs.
Lastly, the translations were filtered for structure retention and for translation quality. On top of SFT, we also created a DPO dataset targeting only specific structured errors that our fine-tuned models exhibit. These included overly copying or parsing issues while ensuring the preferred completions remain as pristine as possible

\subsubsection{Terminology Translation}

Terminology translation is the task of producing a translation that adheres to a glossary of source-target term pairs supplied in the prompt. When a customer has an established vocabulary, such as product names, brand names or domain-specific jargon, it is critical that a translation solution render these consistently and have the ability to use translation memories. We focus on building a preference dataset for this capability. 

For each document, we prompt an LLM to identify the named entities in the source together with their renderings in the reference translation, to propose an alternative but equally valid rendering for each of them in the target language, and to rewrite the reference translation so that it uses these new renderings while leaving the rest of the text untouched. The rewritten translation then becomes the preferred completion and the original reference the dis-preferred one, with the newly proposed term pairs given to the model as the glossary. Since the two completions differ only in how the terms are rendered, the preference signal isolates terminology adherence from the general translation quality.

Notably, a model that blindly copies every glossary entry into its output would score well on such data.  To counterbalance this, we populate the glossary with negative entries for half the documents: source-language words and phrases that are topically related to the document but do \textit{not} occur in it. These are proposed by an LLM, filtered with a substring check against the source and both completions, and verified by a second LLM pass that also rejects inflected forms, compounds and transliterations. A document never receives more negative entries than genuine ones. The resulting dataset contains preference pairs from English into ten languages: Arabic, German, Spanish, French, Hindi, Italian, Japanese, Korean, Portuguese and Russian. Following the strategy described in \Cref{subsec:diversity}, the glossary itself is presented in a randomized format, either as a JSON object, a JSON list of pairs, or one term pair per line with a varying separator, and is attached to a randomly selected instruction. Whenever negative entries are present, the instruction is drawn from a separate set that warns the model that some entries may not be relevant to the source text.

\subsubsection{Tone and Idiomatic Translation}

Human evaluation in early stages revealed that our model struggled with adhering to an expected tone, formality levels and usage of idioms.

To improve the model's adherence to formality instructions, we collected social-domain data in eight languages with explicit formality distinctions: Dutch, Italian, Polish, Greek, Japanese, Spanish, Korean, and Portuguese. Each sample was automatically assigned one of three labels using LLM: \emph{formal}, \emph{informal}, or \emph{ambivalent}. Both grammatical markers and stylistic cues were considered when determining the appropriate formality label.

We then back-translated each target-language sample, into English using our internal MT model, producing supervised training triplets of source back-translated English segment, target language with authentic formality level and associated formality level.
During training, each label was mapped to its corresponding natural-language instruction in the model prompt highlighting the expected formality level.

In order to tackle the idiom translation, we created multilingual synthetic datasets from existing English monolingual idiom datasets. The collected trainset contained 50k idiomatic sentences paired with annotated idioms or literal paraphrases. 

For samples with literal paraphrases, we took them as the source texts for translation and translated them from English into 35 languages using our internal MT model. We then scored each translation with a GEMBA-ESA \citep{kocmi-federmann-2023-large} with a strong LLM as the judge. We retained only examples with a score of at least 70 points. 
For idioms without the literal paraphrases, we adopted a \texttt{best-of-N} approach where we translated each with a mix of several models and selecting the best candidate using the LLM judge. Furthermore, we followed the same approach as above. Finally, we took the top 10k candidates per language pair creating a dataset of 512k sentences.

\subsubsection{Error Detection and Quality Estimation}

Error detection is the task of identifying erroneous spans in a given machine-translated text, while quality estimation assessing the quality of the translated text. Both are evaluated annually by the WMT Automatic Evaluation task \citep{lavie-etal-2025-findings}. 

In our preliminary experiments, training on a fully human labels alone was only marginally better than the baseline, lagging far behind state-of-the-art models, which may be caused by inconsistency in the human annotations.
We therefore focus on synthetic annotations of actual MT errors only, using a strong judge to assess error spans and scores. 

Inspired by \citet{treviso-etal-2024-xtower}, we used an XML tag scheme to annotate the text. We designed a unified XML tag scheme that can represent both: ESA-style tags~\citep{kocmi-etal-2024-error} encoding only severity (\texttt{<error severity="minor or major">}), and MQM-style tags~\citep{lommel2014multidimensional} that additionally carry an error category and subcategory (\texttt{<error severity="..." category="accuracy/mistranslation">}).

\subsubsection{Post-editing}

Post-editing is the process of improving existing translation to capture inaccuracies in fluency and adequacy. Typically, one provides a tuple containing the source and raw translation, and expects an improved translation from an oracle. The oracle can be a human or an automated translation system or model. In our setup, we use another capable judge LLM to be an oracle.

WMT APE  came up with the formulation for Automatic Post-editing, where  people were given human post-edits of some unknown MT system and were allowed to train ML models that can perform similar fixes - when trained on supervised datasets comprising of the original translation, source text and the corrected output \citep{bojar-etal-2015-findings,chatterjee-etal-2018-findings}. 
As human-based post-edits are scarce and harder to scale, people have relied on synthetically generated edits made by LLMs ~\citep{bang-etal-2023-multitask,ki-carpuat-2024-guiding,singh-etal-2025-quality}. Furthermore, one can fine-tune the LLM to imbibe more faithful post-editing capabilities. We adopt this approach, and believe that the ability to post-edit responses at inference time a useful trait for translation models.

We use post-editing both to refine training targets and to teach our model to repair drafts at inference-time. To generate our post-training dataset, we use a base internal MT system that produces an initial translation; which is then passed on to a stronger LLM that post-edits it. We score the draft and the revision with reference-free GEMBA-ESA~\citep{kocmi-federmann-2023-large} and keep the pair only when the post-edit improves the draft, restricting supervision to major quality gaps. Each retained example is packed in two instruction formats: a)~\emph{Blind} post-editing, that provides the source and the raw translation as input, and b)~\emph{Span-assisted} post-editing, which marks erroneous spans in the original translation with similar XML severity tags (major or minor), and provides this annotated translation together with the source as input \citep{lommel2014multidimensional,kocmi-etal-2024-error}.

\subsubsection{Instruction Diversity}
\label{subsec:diversity}

Standard approaches to instruction-tuned MT often rely on static, hardcoded prompts (e.g., "Translate this to [Language]"). While functional, we found out fixed phrasing causes overfitting to the formulation. To enhance robustness of instruction-following capabilities, we replace static prompt instructions with a highly diversified instruction strategy during the training phase.

We introduce variance across three distinct tiers. First, we seed instructions with a curated set of manually prepared paraphrases to capture human-written translation requests. Second, we scale by automatically preparing thousands of instruction variants through paraphrasing, ensuring broad lexical and syntactic coverage. Finally, we translate each variant into all supported languages. When preparing the training dataset, we randomly select a prompt alternating the instruction language between English, the source language, and the target language.
Furthermore, we added various augmentations such as merging consecutive sentences into a single line, replacing new lines with random paragraph separators, enclosing the source and/or target text into XML tags or markdown.

To strengthen instruction adherence, we included a small preference dataset in the last step of DPO. We want the model to follow specified terminology and formality levels, but also to respect custom formatting or style constraints and potentially several constraints at once. We build a dataset synthetically by using an LLM to add creative translation guidelines to existing translation prompts, then generating translations with best-of-N approach selecting best translation with a judge which specifically tests the guideline adherence. The worst completion is then used as a negative example.

\subsection{Reliable Evaluation Metric}
\label{ssec:llm_as_a_judge}

\begin{table}
\centering
\begin{tabular}{lc}
\hline
Metric & Acc (ties) \\
\hline
GEMBA-ESA Fluency+Adequacy & 51.0 \\
GEMBA-ESA & 50.6 \\
xCOMETxl       & 44.9 \\
spBLEU         & 43.7 \\
\hline
\end{tabular}
\caption{Metric meta-evaluation comparing LLM-as-a-judge (with GPT 5.4) versus traditional metrics on internal human label translations}
\label{tab:metric-accuracy}
\end{table}

Determining reliable evaluation metrics was crucial for our model development. We conducted a meta-evaluation using internal human annotations for 24k MT translated segments across 10 languages, applying the segment-level pairwise accuracy methodology introduced in \citet{deutsch-etal-2023-ties}. As shown in \Cref{tab:metric-accuracy}, our results confirm the \citet{lavie-etal-2025-findings}: LLM-as-a-Judge paradigms significantly outperform traditional string-based metrics like xComet or spBLEU. Consequently, we focused on LLM-as-a-judge across our work. 
Furthermore, we designed an updated variant of the GEMBA-ESA \citep{kocmi-federmann-2023-large} prompt to evaluate fluency and adequacy separately before averaging the scores. This decoupled approach yields slightly higher correlation with human judgments while providing the analytical flexibility to diagnose translation accuracy and fluency independently, as well as individual error annotations. The full prompt that we used across the work with various judges is in \Cref{app:gemba-esa-fluade}.

\subsection{Agentic Translation}

To maximize inference throughput, we train a non-reasoning model. While this design choice significantly improves generation speed, it limits the potential quality gains that the model might achieve through intermediate reasoning steps. To bridge this gap for quality-critical applications, we implemented a complementary framework which we refer to as \emph{Agentic Translation}. This approach orchestrates a multi-step workflow by utilizing the exact same underlying model across three distinct functional roles. Initially, a translation agent processes the source text to generate a baseline draft. Subsequently, an evaluation agent inspects this draft to identify and annotate any translation errors. Finally, a post-editing agent leverages the identified error spans to refine and correct the draft, producing the final output. By decomposing the translation process into iterative, targeted steps, the agentic approach recovers the performance benefits typically associated with reasoning models.

\section{Evaluation and Results}

We analyze the performance of our model over MT capabilities and related tasks. We compare it our previous models, top-performing MT open-weight systems, and leading providers as Google Translate and DeepL Pro (Full list in \Cref{sec:benchmark-models}). 
We evaluate all systems in an identical setup, in a clean zero-shot approach without any inference techniques. 

\subsection{Benchmark Models}
\label{sec:benchmark-models}

Next to our new model, we evaluate Cohere's two previous models: Command A Translate \citep{kocmi-etal-2025-command} and general purpose Command A+ \citep{cohere2026commandaplus}. We compare them with a broad selection of top performing open-weight models below 1T total parameters: Google's Gemma~4 31B~\citep{googleaiedge2026gemma4}; Mistral Large~3~\citep{mistralai2025mistral3} and Mistral Medium~3.5~\citep{mistralai2026mistralmedium35}; Alibaba's Qwen3.5 397B~\citep{qwen2026qwen35} and Qwen3.6 27B~\citep{qwen2026qwen36}; Z.ai's GLM-5.2~\citep{glm5team2026glm5vibecodingagentic}; OpenAI's GPT-OSS 120B~\citep{openai2025gptoss120bgptoss20bmodel}; NVIDIA's Nemotron~3 Ultra~\citep{nvidia2026nemotron3ultra}; Thinking Machines Lab's Inkling-Small~\citep{thinkingmachines2026inklingsmall}; and Meta's Muse Glimmer 30B~\citep{cuenca2026museglimmer}. In addition, we compare against two leading commercial translation services: Google Translate and DeepL's NextGen.

We keep the reasoning on for every model that supports reasoning capabilities: specifically, Command A+, Gemma~4 31B, Qwen3.5-397B, Qwen3.6-27B, GLM-5.2, GPT-OSS 120B, and Muse Glimmer 30B.

\subsection{Machine Translation Capabilities}

Our core focus is MT quality across all supported languages. To honestly evaluate capabilities, we evaluate them on a testset released after we finished building the model: the WMT 2026 General MT test set \citep{kocmi-etal-2026-findings}. This dataset contains four distinct domains and translation modes. 

\citet{lavie-etal-2025-findings} showed that traditional evaluation techniques such as xComet \citep{guerreiro-etal-2024-xcomet} or MetricX \citep{juraska-etal-2024-metricx} no longer correlate well with human evaluation and instead strong frontier models excel at evaluation. We observed the same on our internal evaluation described in \Cref{ssec:llm_as_a_judge}.
Thus, we use GPT-5.6 Sol judge without references for evaluation, which we ask to separately judge adequacy and fluency. As a final score we average both scores.

WMT26 is a document-level test set, where number of documents differ for each domain. Building on the recommendation of the original test set, we weight each score by the number of segments in the document. This way each domain is roughly equally represented with about 180 segments.

The speech domain test set is build to test multimodal capabilities. However, all reported models are evaluated on a text-to-text basis. To obtain a source text, we use the provided automatic speech transcription. Finally, we provide the LLM judge the human-cleaned transcript to reduce noise from the ASR.

Results are in \Cref{tab:language_scripts}. \nst scores the top performance across the set of all languages, especially highlighting our strength in European and Latin-script based languages. It reaches second-best performance over languages of the Middle-East and Asia.
Performance over each language is shown in the \Cref{app:per_language}.

\begin{table*}[t]
\centering
\resizebox{\textwidth}{!}{%
\begin{tabular}{llrrrrr}
\toprule
 & \makecell{Params\\total (active)} & \makecell{All\\languages} & \makecell{Europe} & \makecell{ME + Asia} & \makecell{Latin\\script} & \makecell{Other\\scripts} \\
\midrule
\cellcolor[HTML]{D9F0D3}North Small Translate (Agentic) & 218B (25B) & {\cellcolor[HTML]{668F76}} 84.4 & {\cellcolor[HTML]{668F76}} 82.5 & {\cellcolor[HTML]{6E9F7F}} 87.2 & {\cellcolor[HTML]{668F76}} 83.0 & {\cellcolor[HTML]{668F76}} 86.5 \\
\cellcolor[HTML]{D9F0D3}North Small Translate & 218B (25B) & {\cellcolor[HTML]{6B997B}} 83.6 & {\cellcolor[HTML]{699579}} 82.0 & {\cellcolor[HTML]{76AD87}} 86.1 & {\cellcolor[HTML]{6A967A}} 82.4 & {\cellcolor[HTML]{6E9D7E}} 85.4 \\
Mistral Large 3 & 675B (41B) & {\cellcolor[HTML]{7EB58C}} 81.6 & {\cellcolor[HTML]{75AC86}} 80.3 & {\cellcolor[HTML]{A0D1A3}} 83.5 & {\cellcolor[HTML]{7BB28A}} 80.3 & {\cellcolor[HTML]{84BA90}} 83.5 \\
Qwen 3.5 397B & 397B (17B) & {\cellcolor[HTML]{80B68D}} 81.6 & {\cellcolor[HTML]{6E9F7F}} 81.2 & {\cellcolor[HTML]{BCE1BB}} 82.0 & {\cellcolor[HTML]{73A884}} 81.0 & {\cellcolor[HTML]{98CB9D}} 82.4 \\
DeepL NextGen &  & {\cellcolor[HTML]{83B98F}} 81.4 & {\cellcolor[HTML]{679178}} 82.3 & {\cellcolor[HTML]{DBF1D7}} 80.0 & {\cellcolor[HTML]{6A967A}} 82.4 & {\cellcolor[HTML]{CBEAC8}} 79.9 \\
Gemma 4 31b & 31B (31B) & {\cellcolor[HTML]{A3D3A6}} 79.5 & {\cellcolor[HTML]{E8F6E5}} 73.5 & {\cellcolor[HTML]{668F76}} 88.4 & {\cellcolor[HTML]{D9F0D5}} 74.9 & {\cellcolor[HTML]{679178}} 86.3 \\
Inkling Small & 276B (12B) & {\cellcolor[HTML]{B7DEB6}} 78.5 & {\cellcolor[HTML]{9ECFA2}} 77.8 & {\cellcolor[HTML]{E1F3DD}} 79.6 & {\cellcolor[HTML]{9ECFA2}} 78.2 & {\cellcolor[HTML]{D9F0D5}} 79.0 \\
Muse Glimmer 30B & 30B (30B) & {\cellcolor[HTML]{BEE3BC}} 78.2 & {\cellcolor[HTML]{B1DBB2}} 76.8 & {\cellcolor[HTML]{D9F0D5}} 80.3 & {\cellcolor[HTML]{B0DAB1}} 77.4 & {\cellcolor[HTML]{D1EDCE}} 79.5 \\
\cellcolor[HTML]{D9F0D3}Command A+ & 218B (25B) & {\cellcolor[HTML]{D9F0D5}} 76.5 & {\cellcolor[HTML]{C3E6C1}} 76.0 & {\cellcolor[HTML]{F5F9F4}} 77.4 & {\cellcolor[HTML]{C9E9C5}} 76.0 & {\cellcolor[HTML]{EDF7EB}} 77.3 \\
GLM 5.2 & 744B (40B) & {\cellcolor[HTML]{D9F0D5}} 76.5 & {\cellcolor[HTML]{A3D3A6}} 77.5 & {\cellcolor[HTML]{F3EBF4}} 75.0 & {\cellcolor[HTML]{A7D5A9}} 77.8 & {\cellcolor[HTML]{F5EFF5}} 74.6 \\
Mistral Medium 3.5 & 128B (128B) & {\cellcolor[HTML]{E5F5E1}} 75.7 & {\cellcolor[HTML]{E6F5E2}} 73.6 & {\cellcolor[HTML]{EAF6E7}} 78.7 & {\cellcolor[HTML]{E9F6E5}} 73.6 & {\cellcolor[HTML]{DBF1D7}} 78.7 \\
GPT-OSS 120B & 117B (5B) & {\cellcolor[HTML]{F6F2F6}} 72.3 & {\cellcolor[HTML]{F6F2F6}} 70.3 & {\cellcolor[HTML]{F4EDF4}} 75.2 & {\cellcolor[HTML]{FAFAFA}} 71.4 & {\cellcolor[HTML]{F1E5F1}} 73.5 \\
\cellcolor[HTML]{D9F0D3}Command A Translate (2025) & 111B (111B) & {\cellcolor[HTML]{E7D9EB}} 70.0 & {\cellcolor[HTML]{EFE3F0}} 68.8 & {\cellcolor[HTML]{D8C6E0}} 71.7 & {\cellcolor[HTML]{F2E9F3}} 69.6 & {\cellcolor[HTML]{D0BCD9}} 70.5 \\
Google Translate &  & {\cellcolor[HTML]{D5C2DE}} 68.2 & {\cellcolor[HTML]{E8DAEB}} 68.0 & {\cellcolor[HTML]{B794C0}} 68.4 & {\cellcolor[HTML]{EFE3F0}} 68.9 & {\cellcolor[HTML]{AC7FB4}} 67.1 \\
Nemotron 3 Ultra 550B & 550B (55B) & {\cellcolor[HTML]{C9B2D3}} 67.1 & {\cellcolor[HTML]{B38CBC}} 62.9 & {\cellcolor[HTML]{E7D9EB}} 73.3 & {\cellcolor[HTML]{C3AACD}} 64.6 & {\cellcolor[HTML]{D3C0DC}} 70.8 \\
Qwen 3.6 27B & 27B (27B) & {\cellcolor[HTML]{8C6693}} 61.9 & {\cellcolor[HTML]{8C6693}} 59.9 & {\cellcolor[HTML]{8C6693}} 64.9 & {\cellcolor[HTML]{8C6693}} 59.9 & {\cellcolor[HTML]{8C6693}} 65.0 \\
\bottomrule
\end{tabular}
}
\caption{Machine translation capabilities aggregated over different groups of languages over the WMT26 testset.}
\label{tab:language_scripts}
\vspace{-1em}
\end{table*}

We supplement results with a well-known evaluation in the MT community: WMT24++ \citep{deutsch-etal-2025-wmt24}.  WMT24++ contains English to 55 human-translated languages and dialects. The original source text is from \citet{kocmi-etal-2024-findings} and covers multiple domains. Each language pair contains 171 documents split into 998 mostly paragraph level segments. We use the prompt instruction from \citet{deutsch-etal-2025-wmt24} and we evaluate with xComet-XL \citep{guerreiro-etal-2024-xcomet} a metric widely used for system rankings. The metric is a 3.5B parameter model fine-tuned on human judgment data.

\Cref{tab:general_results} shows that \nst reaches near-top performance. That said, caution is needed; as discussed in the \Cref{ssec:saturated_testsets}, WMT24++ is significantly easier than WMT26, and mostly ``solved'' with top-performing systems.

\subsection{Long-Context Translation}
\label{sec:long-context-mt}

We evaluate long-context translation using a subset of the WMT25 General MT test set \citep{kocmi-etal-2025-findings}. The English source long-context comprises of two documents with approximately 9,900 words divided into two sections. We concatenate them into a single source document, yielding one long-form evaluation instance. As part of the evaluation, we translate it into Egyptian Arabic, Czech, Estonian, Icelandic, Japanese, Korean, Russian, Simplified Chinese, and Ukrainian. The complete document is provided to the model in a single request.

Translations are evaluated against the WMT25 references using the reference-based xCOMET-XL metric \citep{guerreiro-etal-2024-xcomet}. Since the metric is not able to evaluate full document, we divide the translations at segment boundaries. xCOMET-XL independently scores each aligned source--hypothesis--reference triple, and returns the arithmetic mean over the 85 segments. 

\Cref{tab:general_results} shows that our model reachs second-best MT quality, outperformed only by DeepL.\footnote{DeepL is a closed-source system accessed by API, so we cannot confirm implementation: for instance, whether it might pre-parse the source text.}

\subsection{Terminology Translation}
\label{sec:terminology-translation}

We evaluate glossary-conditioned translation using the WMT25 Terminology Translation Task \citep{semenov-etal-2025-findings}. We refer readers to that work for the dataset construction, domains, and official shared-task protocol. Our evaluation covers five translation directions: English to German, Spanish, Russian, and Traditional Chinese, and Traditional Chinese to English.

For the scoring we use only the proper-terminology condition setup. Each model receives the source text together with a source--target glossary and is instructed to apply the supplied terminology while returning only the translation.

We measure is corpus-level terminology success rate: a dictionary pair is successful when its source term occurs in the input and its prescribed target form occurs in the translation. As a final score in \Cref{tab:general_results}, we micro average language pair scores from both WMT25 Terminology Tracks.

\subsection{Evaluation of Structured Translation}

This task is a first of its kind, hence efforts were dedicated towards building test suites as well as designing relevant metrics for this task. 
Our test suites were developed using internal datasets, from the same distribution of training data and also naturally occurring sentences that were combined to create diverse JSON structures. The examples included two translation tasks: translating entire structures with all fields, and translating only user-specified fields, designed to evaluate the model's instruction-following abilities. 
We also included a markdown evaluation suite but our base models were already saturating their performance, thus we subsequently dropped this test suite. 

For evaluation of structured translation, we tracked two areas - the quality of translation and primarily the consistency of the structured translation, which we report in this paper. The evaluation has multiple metrics targeting various errors observed during model development. We report an average score of all the metrics described below in \cref{tab:general_results}.
\begin{itemize}
    \item Overall structure: Three metrics - \textit{parsing accuracy} checks if the produced translation adheres to the right delimiters from which the structured output can be extracted. \textit{Decoding accuracy} determines if the extracted output can be decoded into a valid object (e.g., a JSON or YAML object). And \textit{schema accuracy} checks whether the decoded translation output matches the source schema (e.g., the same JSON keys). These metrics have cascading effect on the next structured metrics.
    \item Untranslation issues: The intermediate models exhibited different levels of copying in their outputs, from leaving intermediate fields untranslated to entire translations retained in the source language. In addition, the instruction following test suite involved only translating a limited number of fields, requiring the remaining fields to be retained in their source language. Thus, the evaluation measured (i) percentage of samples where all non translatable content was copied correctly and (ii) average per-sample percentage of translatable fields that were incorrectly copied.
    Lastly, a separate metric evaluated the proportion of examples that were left completely untranslated. 
    \item Line count consistency: This metric operates on a per-field level and checked if the lines in generated translation match exactly the number of lines in source translation.
\end{itemize}

\subsection{Quantifying Test Set Saturation}
\label{ssec:saturated_testsets}

To verify the headroom for improvement for popular WMT testsets, we ran a human evaluation using the same system and the ESA protocol \citep{kocmi-etal-2024-error}. \Cref{tab:major-errors-by-year} showx that for high-resource languages, the WMT24 testset is much easier than WMT26---nearly solved. This is especially problematic in combination with xComet where the score differences are more affected by the metric word preferences rather than the actual errors, which are getting rare. Furthermore, we observed that for the WMT 2026 the software domain is the easiest domain, while other domains remain considerably more challenging.

These results should be interpreted with some caution. Although both evaluations were conducted using the Pearmut framework \citep{zouhar2026pearmut}, they differ in certain evaluation details, most notably affecting the slider anchors. We argue, however, that these changes had minimal impact on the major error annotation, which remained comparable. A second difference is that each testset was evaluated in a different month, alongside a distinct set of systems, and by a slightly different pool of annotators. While the annotator pool could not be strictly controlled, we selected one system that was identical across both evaluations. Thus, the table shows the percentage of errors for this specific model across both setups. For context, this is a strong system that outperforms Command A+, though it underperforms North Small Translate.

\begin{table}[t]
\centering
\small
\begin{tabular}{lrr}
\toprule
Language & WMT24 & WMT26 \\
\midrule
Arabic (MSA)         & 19.7\% & 34.9\% \\
German               & 22.9\% & 37.1\% \\
Spanish (Mexico)     & 14.2\% & 34.9\% \\
French               & 7.6\%  & 13.9\% \\
Italian              & 11.9\% & 11.5\% \\
Japanese             & 14.6\% & 28.9\% \\
Korean               & 29.9\% & 19.3\% \\
Portuguese (Brazil)  & 6.6\%  & 17.5\% \\
Chinese (Simplified) & 14.4\% & 18.1\% \\
\midrule
\textbf{Overall difficulty} & \textbf{15.8\%} & \textbf{24.0\%} \\
\bottomrule
\end{tabular}
\caption{Percentage of segments containing at least one major error. Highlighting the saturation of the test sets.}
\label{tab:major-errors-by-year}
\end{table}

\section{Conclusion}
We introduce \textit{\nst}, Cohere's state-of-the-art machine translation system. A sparse mixture-of-experts non-reasoning transformer-based LLM with 25 billion active and 218 billion total parameters, it achieves top-tier performance across 50 languages. 

We build \textit{\nst} via a five-step training protocol spanning SFT, DPO, and online RL. By using difficulty sampling and an orchestrated preference distillation pipeline, we focused our synthetic data generation and training signal strictly on the most challenging documents. Consequently, the model excels in standard machine translation as well as complex related tasks, including structured translation, terminology adherence, and error detection.

For latency-insensitive applications requiring maximum quality, we incorporated an Agentic Translation framework, enabling the model to iteratively evaluate and post-edit its own drafts. We submitted our model to the WMT 2026 General MT shared task as \textit{Cohere CAT+}, demonstrating the model's strong out-of-the-box capabilities.

\section*{Limitations}

The evaluation of MT systems is fundamentally limited by the noise and limited discriminative power of automated benchmarks, and even of human evaluators. Translation quality can be subjective, and furthermore, high translation quality in one domain for a given language does not guarantee high quality in another, even for the same language. Preferred system recommendations can thus change depending on use case.

\bibliography{custom,anthology-1,anthology-2}

\appendix

\begin{table*}[t]
\centering
\resizebox{\textwidth}{!}{%
\begin{minipage}{1.7\textwidth}
\begin{tabular}{llrrrrrrrrrrrr}
\toprule
 & \makecell{Params\\total (active)} & \makecell{All\\languages} & \makecell{Albanian} & \makecell{Bengali} & \makecell{Bulgarian} & \makecell{Catalan} & \makecell{Croatian} & \makecell{Czech} & \makecell{Danish} & \makecell{Dutch} & \makecell{Egyptian\\Arabic} & \makecell{Estonian} & \makecell{European\\Portuguese} \\
\midrule
\cellcolor[HTML]{D9F0D3}North Small Translate (Agentic) & 218B (25B) & {\cellcolor[HTML]{668F76}} 84.4 & {\cellcolor[HTML]{6E9D7E}} 78.7 & {\cellcolor[HTML]{6E9F7F}} 87.5 & {\cellcolor[HTML]{93C79A}} 80.1 & {\cellcolor[HTML]{6C9B7D}} 84.0 & {\cellcolor[HTML]{C3E6C1}} 79.0 & {\cellcolor[HTML]{699579}} 88.8 & {\cellcolor[HTML]{668F76}} 89.1 & {\cellcolor[HTML]{668F76}} 89.5 & {\cellcolor[HTML]{75AC86}} 84.2 & {\cellcolor[HTML]{ACD8AE}} 73.5 & {\cellcolor[HTML]{668F76}} 90.4 \\
\cellcolor[HTML]{D9F0D3}North Small Translate & 218B (25B) & {\cellcolor[HTML]{6B997B}} 83.6 & {\cellcolor[HTML]{70A281}} 77.9 & {\cellcolor[HTML]{ACD8AE}} 82.8 & {\cellcolor[HTML]{9CCEA0}} 79.5 & {\cellcolor[HTML]{6D9C7D}} 83.9 & {\cellcolor[HTML]{CDEBC9}} 78.3 & {\cellcolor[HTML]{668F76}} 89.2 & {\cellcolor[HTML]{A0D1A3}} 85.3 & {\cellcolor[HTML]{668F76}} 89.5 & {\cellcolor[HTML]{76AD87}} 84.0 & {\cellcolor[HTML]{B9DFB8}} 72.6 & {\cellcolor[HTML]{699579}} 89.6 \\
Mistral Large 3 & 675B (41B) & {\cellcolor[HTML]{7EB58C}} 81.6 & {\cellcolor[HTML]{75AC86}} 76.2 & {\cellcolor[HTML]{8ABF94}} 84.9 & {\cellcolor[HTML]{6E9F7F}} 83.2 & {\cellcolor[HTML]{668F76}} 85.3 & {\cellcolor[HTML]{98CB9D}} 81.4 & {\cellcolor[HTML]{A5D4A8}} 83.9 & {\cellcolor[HTML]{90C498}} 86.0 & {\cellcolor[HTML]{76AD87}} 87.9 & {\cellcolor[HTML]{B5DDB5}} 78.7 & {\cellcolor[HTML]{C2E5BF}} 72.0 & {\cellcolor[HTML]{77AF87}} 86.4 \\
Qwen 3.5 397B & 397B (17B) & {\cellcolor[HTML]{80B68D}} 81.6 & {\cellcolor[HTML]{668F76}} 81.2 & {\cellcolor[HTML]{9BCD9F}} 83.8 & {\cellcolor[HTML]{6E9D7E}} 83.3 & {\cellcolor[HTML]{76AD87}} 82.0 & {\cellcolor[HTML]{7AB189}} 83.4 & {\cellcolor[HTML]{D0ECCD}} 81.5 & {\cellcolor[HTML]{9BCD9F}} 85.5 & {\cellcolor[HTML]{89BE93}} 87.1 & {\cellcolor[HTML]{D9F0D5}} 75.4 & {\cellcolor[HTML]{90C498}} 75.6 & {\cellcolor[HTML]{B3DCB3}} 80.7 \\
DeepL NextGen &  & {\cellcolor[HTML]{83B98F}} 81.4 & {\cellcolor[HTML]{75AB85}} 76.5 & {\cellcolor[HTML]{F9F6F9}} 75.8 & {\cellcolor[HTML]{668F76}} 84.5 & {\cellcolor[HTML]{9CCEA0}} 78.7 & {\cellcolor[HTML]{E3F4E0}} 76.4 & {\cellcolor[HTML]{8DC296}} 85.3 & {\cellcolor[HTML]{6A977B}} 88.6 & {\cellcolor[HTML]{668F76}} 89.5 & {\cellcolor[HTML]{B48DBD}} 57.6 & {\cellcolor[HTML]{668F76}} 81.0 & {\cellcolor[HTML]{73A884}} 87.2 \\
Gemma 4 31b & 31B (31B) & {\cellcolor[HTML]{A3D3A6}} 79.5 & {\cellcolor[HTML]{76AD87}} 76.2 & {\cellcolor[HTML]{668F76}} 88.8 & {\cellcolor[HTML]{B0DAB1}} 78.5 & {\cellcolor[HTML]{A1D2A5}} 78.3 & {\cellcolor[HTML]{BCE1BB}} 79.4 & {\cellcolor[HTML]{F4ECF4}} 75.5 & {\cellcolor[HTML]{6B997B}} 88.5 & {\cellcolor[HTML]{6C9A7C}} 88.9 & {\cellcolor[HTML]{668F76}} 87.5 & {\cellcolor[HTML]{B188BA}} 53.1 & {\cellcolor[HTML]{74AA85}} 87.0 \\
Inkling Small & 276B (12B) & {\cellcolor[HTML]{B7DEB6}} 78.5 & {\cellcolor[HTML]{87BD92}} 73.8 & {\cellcolor[HTML]{D7EFD4}} 80.2 & {\cellcolor[HTML]{B1DBB2}} 78.4 & {\cellcolor[HTML]{74AA85}} 82.4 & {\cellcolor[HTML]{E0F3DC}} 76.7 & {\cellcolor[HTML]{EEF7EB}} 78.8 & {\cellcolor[HTML]{DBF1D7}} 82.5 & {\cellcolor[HTML]{DAF0D6}} 83.9 & {\cellcolor[HTML]{A1D2A5}} 80.1 & {\cellcolor[HTML]{B1DBB2}} 73.1 & {\cellcolor[HTML]{AED9AF}} 81.1 \\
Muse Glimmer 30B & 30B (30B) & {\cellcolor[HTML]{BEE3BC}} 78.2 & {\cellcolor[HTML]{77AF87}} 76.0 & {\cellcolor[HTML]{B3DCB3}} 82.4 & {\cellcolor[HTML]{CBEAC8}} 76.9 & {\cellcolor[HTML]{87BD92}} 80.5 & {\cellcolor[HTML]{DCF1D9}} 77.0 & {\cellcolor[HTML]{ECF7EA}} 79.1 & {\cellcolor[HTML]{F9F8F9}} 79.6 & {\cellcolor[HTML]{F5F9F4}} 82.0 & {\cellcolor[HTML]{90C498}} 81.6 & {\cellcolor[HTML]{A7D5A9}} 73.8 & {\cellcolor[HTML]{98CB9D}} 83.1 \\
\cellcolor[HTML]{D9F0D3}Command A+ & 218B (25B) & {\cellcolor[HTML]{D9F0D5}} 76.5 & {\cellcolor[HTML]{8ABF94}} 73.4 & {\cellcolor[HTML]{E2F4DF}} 79.3 & {\cellcolor[HTML]{DFF2DB}} 75.4 & {\cellcolor[HTML]{9BCD9F}} 78.8 & {\cellcolor[HTML]{E8F6E5}} 76.0 & {\cellcolor[HTML]{B0DAB1}} 83.3 & {\cellcolor[HTML]{A5D4A8}} 85.0 & {\cellcolor[HTML]{EBF7E8}} 82.9 & {\cellcolor[HTML]{F2F8F0}} 71.8 & {\cellcolor[HTML]{F6F9F5}} 65.9 & {\cellcolor[HTML]{C0E4BE}} 79.5 \\
GLM 5.2 & 744B (40B) & {\cellcolor[HTML]{D9F0D5}} 76.5 & {\cellcolor[HTML]{6D9C7D}} 79.0 & {\cellcolor[HTML]{E7D9EB}} 72.6 & {\cellcolor[HTML]{B5DDB5}} 78.2 & {\cellcolor[HTML]{6C9A7C}} 84.1 & {\cellcolor[HTML]{668F76}} 86.2 & {\cellcolor[HTML]{7EB58C}} 86.3 & {\cellcolor[HTML]{A7D5A9}} 85.0 & {\cellcolor[HTML]{F6F1F6}} 80.9 & {\cellcolor[HTML]{E7F6E3}} 73.8 & {\cellcolor[HTML]{7BB28A}} 77.3 & {\cellcolor[HTML]{F7F2F7}} 69.5 \\
Mistral Medium 3.5 & 128B (128B) & {\cellcolor[HTML]{E5F5E1}} 75.7 & {\cellcolor[HTML]{EAF6E6}} 60.4 & {\cellcolor[HTML]{CEEBCA}} 80.9 & {\cellcolor[HTML]{B1DBB2}} 78.4 & {\cellcolor[HTML]{6E9F7F}} 83.6 & {\cellcolor[HTML]{F6F9F6}} 74.1 & {\cellcolor[HTML]{E2F4DF}} 80.1 & {\cellcolor[HTML]{D3EDCF}} 83.0 & {\cellcolor[HTML]{C3E6C1}} 85.0 & {\cellcolor[HTML]{F3EBF4}} 67.7 & {\cellcolor[HTML]{EBDFEE}} 61.1 & {\cellcolor[HTML]{8CC095}} 84.3 \\
GPT-OSS 120B & 117B (5B) & {\cellcolor[HTML]{F6F2F6}} 72.3 & {\cellcolor[HTML]{C3E6C1}} 66.4 & {\cellcolor[HTML]{EFE3F0}} 73.5 & {\cellcolor[HTML]{F5EFF5}} 70.8 & {\cellcolor[HTML]{DCF1D9}} 73.4 & {\cellcolor[HTML]{F4EDF4}} 71.8 & {\cellcolor[HTML]{D2BEDB}} 71.6 & {\cellcolor[HTML]{EADCEC}} 77.4 & {\cellcolor[HTML]{D5C2DE}} 78.0 & {\cellcolor[HTML]{DCF1D9}} 75.0 & {\cellcolor[HTML]{F2E9F3}} 62.5 & {\cellcolor[HTML]{C9E9C5}} 78.8 \\
\cellcolor[HTML]{D9F0D3}Command A Translate (2025) & 111B (111B) & {\cellcolor[HTML]{E7D9EB}} 70.0 & {\cellcolor[HTML]{FAFAFA}} 55.7 & {\cellcolor[HTML]{8C6693}} 63.7 & {\cellcolor[HTML]{8C6693}} 59.7 & {\cellcolor[HTML]{B1DBB2}} 77.2 & {\cellcolor[HTML]{C5ADCF}} 66.3 & {\cellcolor[HTML]{D5EED1}} 81.1 & {\cellcolor[HTML]{D0BCD9}} 75.3 & {\cellcolor[HTML]{EBF7E8}} 82.9 & {\cellcolor[HTML]{8FC397}} 81.7 & {\cellcolor[HTML]{9E74A5}} 51.1 & {\cellcolor[HTML]{93C79A}} 83.5 \\
Google Translate &  & {\cellcolor[HTML]{D5C2DE}} 68.2 & {\cellcolor[HTML]{B9DFB8}} 67.7 & {\cellcolor[HTML]{E9DBEC}} 72.9 & {\cellcolor[HTML]{E2D2E7}} 68.0 & {\cellcolor[HTML]{F5F9F4}} 69.8 & {\cellcolor[HTML]{E1D1E7}} 69.2 & {\cellcolor[HTML]{8F6896}} 65.4 & {\cellcolor[HTML]{916A98}} 70.8 & {\cellcolor[HTML]{C3AACD}} 76.8 & {\cellcolor[HTML]{8C6693}} 53.0 & {\cellcolor[HTML]{F8FAF7}} 65.6 & {\cellcolor[HTML]{8C6693}} 51.6 \\
Nemotron 3 Ultra 550B & 550B (55B) & {\cellcolor[HTML]{C9B2D3}} 67.1 & {\cellcolor[HTML]{F4EDF5}} 52.5 & {\cellcolor[HTML]{A77BAF}} 65.8 & {\cellcolor[HTML]{A176A8}} 61.3 & {\cellcolor[HTML]{D1EDCE}} 74.6 & {\cellcolor[HTML]{8C6693}} 60.8 & {\cellcolor[HTML]{8C6693}} 65.1 & {\cellcolor[HTML]{8C6693}} 70.5 & {\cellcolor[HTML]{8C6693}} 73.6 & {\cellcolor[HTML]{AAD7AC}} 79.5 & {\cellcolor[HTML]{8C6693}} 49.2 & {\cellcolor[HTML]{80B68D}} 85.4 \\
Qwen 3.6 27B & 27B (27B) & {\cellcolor[HTML]{8C6693}} 61.9 & {\cellcolor[HTML]{8C6693}} 30.1 & {\cellcolor[HTML]{BFA3C9}} 68.4 & {\cellcolor[HTML]{C2A9CD}} 64.7 & {\cellcolor[HTML]{8C6693}} 52.2 & {\cellcolor[HTML]{D4EED0}} 77.7 & {\cellcolor[HTML]{C2A9CD}} 70.0 & {\cellcolor[HTML]{F6F1F6}} 79.0 & {\cellcolor[HTML]{F0E4F1}} 79.9 & {\cellcolor[HTML]{DDCCE4}} 63.6 & {\cellcolor[HTML]{AB7EB3}} 52.2 & {\cellcolor[HTML]{96C99C}} 83.2 \\
\bottomrule
\end{tabular}

\begin{tabular}{lrrrrrrrrrrrrr}
\toprule
 & \makecell{Filipino} & \makecell{Finnish} & \makecell{French} & \makecell{German} & \makecell{Greek} & \makecell{Hebrew} & \makecell{Hindi} & \makecell{Hungarian} & \makecell{Icelandic} & \makecell{Indonesian} & \makecell{Irish} & \makecell{Italian} & \makecell{Japanese} \\
\midrule
\cellcolor[HTML]{D9F0D3}North Small Translate (Agentic) & {\cellcolor[HTML]{75AB85}} 82.8 & {\cellcolor[HTML]{699479}} 86.1 & {\cellcolor[HTML]{699479}} 90.2 & {\cellcolor[HTML]{668F76}} 89.7 & {\cellcolor[HTML]{668F76}} 89.4 & {\cellcolor[HTML]{668F76}} 87.1 & {\cellcolor[HTML]{6C9B7D}} 88.8 & {\cellcolor[HTML]{C5E7C2}} 74.8 & {\cellcolor[HTML]{75AB85}} 67.2 & {\cellcolor[HTML]{668F76}} 90.1 & {\cellcolor[HTML]{699479}} 69.6 & {\cellcolor[HTML]{668F76}} 90.3 & {\cellcolor[HTML]{8ABF94}} 87.9 \\
\cellcolor[HTML]{D9F0D3}North Small Translate & {\cellcolor[HTML]{75AC86}} 82.7 & {\cellcolor[HTML]{668F76}} 86.6 & {\cellcolor[HTML]{699579}} 90.2 & {\cellcolor[HTML]{668F76}} 89.7 & {\cellcolor[HTML]{689278}} 89.0 & {\cellcolor[HTML]{6FA07F}} 83.3 & {\cellcolor[HTML]{71A481}} 88.3 & {\cellcolor[HTML]{D1EDCE}} 73.8 & {\cellcolor[HTML]{72A683}} 67.9 & {\cellcolor[HTML]{D0ECCD}} 85.8 & {\cellcolor[HTML]{6C9B7D}} 68.4 & {\cellcolor[HTML]{679077}} 90.2 & {\cellcolor[HTML]{95C89B}} 87.4 \\
Mistral Large 3 & {\cellcolor[HTML]{AED9AF}} 77.5 & {\cellcolor[HTML]{A7D5A9}} 80.0 & {\cellcolor[HTML]{84BA90}} 88.7 & {\cellcolor[HTML]{84BA90}} 87.7 & {\cellcolor[HTML]{BAE0B9}} 81.1 & {\cellcolor[HTML]{75AC86}} 80.6 & {\cellcolor[HTML]{BEE3BC}} 84.6 & {\cellcolor[HTML]{6E9D7E}} 81.4 & {\cellcolor[HTML]{6E9D7E}} 69.1 & {\cellcolor[HTML]{C2E5BF}} 86.3 & {\cellcolor[HTML]{87BD92}} 62.6 & {\cellcolor[HTML]{92C699}} 88.0 & {\cellcolor[HTML]{A3D3A6}} 86.7 \\
Qwen 3.5 397B & {\cellcolor[HTML]{FAF9FA}} 67.9 & {\cellcolor[HTML]{7AB189}} 83.4 & {\cellcolor[HTML]{D9C7E1}} 81.9 & {\cellcolor[HTML]{96C99C}} 87.0 & {\cellcolor[HTML]{CEEBCA}} 79.7 & {\cellcolor[HTML]{71A582}} 82.2 & {\cellcolor[HTML]{F5EEF5}} 79.6 & {\cellcolor[HTML]{668F76}} 82.7 & {\cellcolor[HTML]{668F76}} 71.4 & {\cellcolor[HTML]{C9E9C5}} 86.1 & {\cellcolor[HTML]{70A281}} 67.1 & {\cellcolor[HTML]{BEE3BC}} 86.5 & {\cellcolor[HTML]{FAF9FA}} 81.2 \\
DeepL NextGen & {\cellcolor[HTML]{C9E9C5}} 75.3 & {\cellcolor[HTML]{6E9F7F}} 85.0 & {\cellcolor[HTML]{CFECCB}} 86.6 & {\cellcolor[HTML]{A9D6AB}} 86.4 & {\cellcolor[HTML]{A3D3A6}} 82.7 & {\cellcolor[HTML]{699579}} 85.7 & {\cellcolor[HTML]{E4D5E8}} 77.8 & {\cellcolor[HTML]{668F76}} 82.7 & {\cellcolor[HTML]{75AB85}} 67.2 & {\cellcolor[HTML]{93C79A}} 87.7 & {\cellcolor[HTML]{668F76}} 70.6 & {\cellcolor[HTML]{86BB91}} 88.4 & {\cellcolor[HTML]{D3EDCF}} 84.6 \\
Gemma 4 31b & {\cellcolor[HTML]{668F76}} 86.1 & {\cellcolor[HTML]{84BA90}} 82.6 & {\cellcolor[HTML]{668F76}} 90.4 & {\cellcolor[HTML]{679178}} 89.6 & {\cellcolor[HTML]{F9FAF9}} 74.1 & {\cellcolor[HTML]{75AB85}} 80.8 & {\cellcolor[HTML]{668F76}} 89.5 & {\cellcolor[HTML]{8C6693}} 55.4 & {\cellcolor[HTML]{E4D6E9}} 41.3 & {\cellcolor[HTML]{679077}} 90.0 & {\cellcolor[HTML]{9970A0}} 18.9 & {\cellcolor[HTML]{76AD87}} 89.0 & {\cellcolor[HTML]{668F76}} 90.9 \\
Inkling Small & {\cellcolor[HTML]{D4EED0}} 74.1 & {\cellcolor[HTML]{C2E5BF}} 78.5 & {\cellcolor[HTML]{DFCFE5}} 82.2 & {\cellcolor[HTML]{F7F4F8}} 82.0 & {\cellcolor[HTML]{D3EDCF}} 79.1 & {\cellcolor[HTML]{7DB48B}} 78.9 & {\cellcolor[HTML]{F1E7F2}} 79.0 & {\cellcolor[HTML]{98CB9D}} 77.5 & {\cellcolor[HTML]{6E9D7E}} 69.2 & {\cellcolor[HTML]{E1D1E7}} 81.2 & {\cellcolor[HTML]{6E9D7E}} 67.9 & {\cellcolor[HTML]{EFF7EC}} 84.4 & {\cellcolor[HTML]{F5EFF5}} 80.4 \\
Muse Glimmer 30B & {\cellcolor[HTML]{BCE1BB}} 76.3 & {\cellcolor[HTML]{DBF1D7}} 76.2 & {\cellcolor[HTML]{F5F0F6}} 83.8 & {\cellcolor[HTML]{E3F4E0}} 84.1 & {\cellcolor[HTML]{E0F3DC}} 77.8 & {\cellcolor[HTML]{87BD92}} 77.2 & {\cellcolor[HTML]{F6F2F6}} 79.9 & {\cellcolor[HTML]{8ABF94}} 78.5 & {\cellcolor[HTML]{7EB58C}} 65.7 & {\cellcolor[HTML]{F6F2F6}} 82.9 & {\cellcolor[HTML]{A1D2A5}} 59.1 & {\cellcolor[HTML]{F5F0F6}} 82.8 & {\cellcolor[HTML]{E1F3DD}} 83.7 \\
\cellcolor[HTML]{D9F0D3}Command A+ & {\cellcolor[HTML]{D7EFD4}} 73.6 & {\cellcolor[HTML]{C5E7C2}} 78.2 & {\cellcolor[HTML]{EAF6E6}} 85.5 & {\cellcolor[HTML]{DBF1D7}} 84.5 & {\cellcolor[HTML]{AED9AF}} 81.9 & {\cellcolor[HTML]{73A783}} 81.6 & {\cellcolor[HTML]{F4ECF4}} 79.4 & {\cellcolor[HTML]{F2F8F0}} 70.3 & {\cellcolor[HTML]{E3F4E0}} 53.7 & {\cellcolor[HTML]{CDB8D7}} 80.1 & {\cellcolor[HTML]{CBEAC8}} 54.2 & {\cellcolor[HTML]{E9F6E5}} 84.8 & {\cellcolor[HTML]{F0E4F1}} 79.4 \\
GLM 5.2 & {\cellcolor[HTML]{D8C6E0}} 60.7 & {\cellcolor[HTML]{F8FAF8}} 72.4 & {\cellcolor[HTML]{78B088}} 89.1 & {\cellcolor[HTML]{B7DEB6}} 85.9 & {\cellcolor[HTML]{8E6794}} 58.5 & {\cellcolor[HTML]{A5D4A8}} 72.0 & {\cellcolor[HTML]{CCB7D6}} 76.1 & {\cellcolor[HTML]{D1EDCE}} 73.8 & {\cellcolor[HTML]{83B98F}} 65.1 & {\cellcolor[HTML]{AC7FB4}} 78.1 & {\cellcolor[HTML]{DAF0D6}} 51.7 & {\cellcolor[HTML]{F1F8F0}} 84.1 & {\cellcolor[HTML]{6D9C7D}} 90.0 \\
Mistral Medium 3.5 & {\cellcolor[HTML]{ECF7EA}} 70.8 & {\cellcolor[HTML]{F6F9F6}} 72.8 & {\cellcolor[HTML]{98CB9D}} 88.1 & {\cellcolor[HTML]{9CCEA0}} 86.8 & {\cellcolor[HTML]{EDF7EB}} 76.1 & {\cellcolor[HTML]{8FC397}} 75.8 & {\cellcolor[HTML]{F3F9F2}} 81.3 & {\cellcolor[HTML]{BEE3BC}} 75.1 & {\cellcolor[HTML]{F8FAF8}} 48.8 & {\cellcolor[HTML]{F4F9F3}} 83.9 & {\cellcolor[HTML]{F6F9F6}} 44.9 & {\cellcolor[HTML]{C7E8C4}} 86.3 & {\cellcolor[HTML]{CEEBCA}} 84.9 \\
GPT-OSS 120B & {\cellcolor[HTML]{CDEBC9}} 75.0 & {\cellcolor[HTML]{F1E6F2}} 69.5 & {\cellcolor[HTML]{E5D7EA}} 82.5 & {\cellcolor[HTML]{D3C0DC}} 79.1 & {\cellcolor[HTML]{DCCBE3}} 67.9 & {\cellcolor[HTML]{C3E6C1}} 67.8 & {\cellcolor[HTML]{D4C1DD}} 76.7 & {\cellcolor[HTML]{F3F8F1}} 70.2 & {\cellcolor[HTML]{D6EFD3}} 55.7 & {\cellcolor[HTML]{CCB7D6}} 80.0 & {\cellcolor[HTML]{DFF2DB}} 50.7 & {\cellcolor[HTML]{C2A9CD}} 79.4 & {\cellcolor[HTML]{F3EAF3}} 79.9 \\
\cellcolor[HTML]{D9F0D3}Command A Translate (2025) & {\cellcolor[HTML]{F1E6F2}} 64.7 & {\cellcolor[HTML]{B794C0}} 62.1 & {\cellcolor[HTML]{F9F7F9}} 84.2 & {\cellcolor[HTML]{EFF7EC}} 83.4 & {\cellcolor[HTML]{C7E8C4}} 80.2 & {\cellcolor[HTML]{7AB189}} 79.6 & {\cellcolor[HTML]{DDCCE4}} 77.2 & {\cellcolor[HTML]{FAF9FA}} 68.9 & {\cellcolor[HTML]{E8DAEB}} 42.1 & {\cellcolor[HTML]{EFE3F0}} 82.0 & {\cellcolor[HTML]{F1E5F1}} 38.4 & {\cellcolor[HTML]{F0F8EE}} 84.3 & {\cellcolor[HTML]{EEE2EF}} 79.2 \\
Google Translate & {\cellcolor[HTML]{F7F3F7}} 66.9 & {\cellcolor[HTML]{DAC8E2}} 66.4 & {\cellcolor[HTML]{936B9A}} 78.5 & {\cellcolor[HTML]{8C6693}} 75.1 & {\cellcolor[HTML]{C5ADCF}} 64.9 & {\cellcolor[HTML]{D3EDCF}} 64.9 & {\cellcolor[HTML]{8C6693}} 71.7 & {\cellcolor[HTML]{E1D1E7}} 64.4 & {\cellcolor[HTML]{8CC095}} 64.2 & {\cellcolor[HTML]{9D73A4}} 77.5 & {\cellcolor[HTML]{77AF87}} 65.0 & {\cellcolor[HTML]{8C6693}} 76.7 & {\cellcolor[HTML]{8C6693}} 71.8 \\
Nemotron 3 Ultra 550B & {\cellcolor[HTML]{F1F8EF}} 70.0 & {\cellcolor[HTML]{8C6693}} 57.8 & {\cellcolor[HTML]{C5E7C2}} 86.9 & {\cellcolor[HTML]{E8F6E5}} 83.9 & {\cellcolor[HTML]{8C6693}} 58.4 & {\cellcolor[HTML]{6C9B7D}} 84.4 & {\cellcolor[HTML]{B9DFB8}} 84.9 & {\cellcolor[HTML]{B38CBC}} 59.0 & {\cellcolor[HTML]{F3EBF4}} 45.2 & {\cellcolor[HTML]{8C6693}} 76.8 & {\cellcolor[HTML]{FAF9FA}} 43.4 & {\cellcolor[HTML]{F4F9F3}} 83.9 & {\cellcolor[HTML]{6E9F7F}} 89.9 \\
Qwen 3.6 27B & {\cellcolor[HTML]{8C6693}} 50.1 & {\cellcolor[HTML]{986F9F}} 58.8 & {\cellcolor[HTML]{8C6693}} 78.3 & {\cellcolor[HTML]{9C72A3}} 75.8 & {\cellcolor[HTML]{976E9E}} 59.4 & {\cellcolor[HTML]{8C6693}} 19.4 & {\cellcolor[HTML]{F6F9F5}} 81.1 & {\cellcolor[HTML]{F8FAF8}} 69.3 & {\cellcolor[HTML]{8C6693}} 25.4 & {\cellcolor[HTML]{F9F6F9}} 83.2 & {\cellcolor[HTML]{8C6693}} 16.8 & {\cellcolor[HTML]{EFF7EC}} 84.4 & {\cellcolor[HTML]{C2A9CD}} 75.6 \\
\bottomrule
\end{tabular}

\begin{tabular}{lrrrrrrrrrrrrr}
\toprule
 & \makecell{Korean} & \makecell{Latvian} & \makecell{Lithuanian} & \makecell{Malay} & \makecell{Maltese} & \makecell{Norwegian\\Bokmal} & \makecell{Panjabi} & \makecell{Persian} & \makecell{Polish} & \makecell{Romanian} & \makecell{Russian} & \makecell{Serbian} & \makecell{Simplified\\Chinese} \\
\midrule
\cellcolor[HTML]{D9F0D3}North Small Translate (Agentic) & {\cellcolor[HTML]{7AB189}} 88.9 & {\cellcolor[HTML]{9BCD9F}} 71.6 & {\cellcolor[HTML]{8ABF94}} 69.6 & {\cellcolor[HTML]{74AA85}} 87.3 & {\cellcolor[HTML]{75AC86}} 65.0 & {\cellcolor[HTML]{668F76}} 90.2 & {\cellcolor[HTML]{679077}} 87.8 & {\cellcolor[HTML]{74AA85}} 87.4 & {\cellcolor[HTML]{73A783}} 87.2 & {\cellcolor[HTML]{668F76}} 88.1 & {\cellcolor[HTML]{6E9F7F}} 88.8 & {\cellcolor[HTML]{87BD92}} 74.9 & {\cellcolor[HTML]{95C89B}} 89.1 \\
\cellcolor[HTML]{D9F0D3}North Small Translate & {\cellcolor[HTML]{BCE1BB}} 84.4 & {\cellcolor[HTML]{9ECFA2}} 71.2 & {\cellcolor[HTML]{90C498}} 68.5 & {\cellcolor[HTML]{76AD87}} 87.0 & {\cellcolor[HTML]{74AA85}} 65.6 & {\cellcolor[HTML]{679077}} 90.0 & {\cellcolor[HTML]{679178}} 87.6 & {\cellcolor[HTML]{77AF87}} 87.0 & {\cellcolor[HTML]{73A783}} 87.2 & {\cellcolor[HTML]{689278}} 87.6 & {\cellcolor[HTML]{75AB85}} 88.1 & {\cellcolor[HTML]{92C699}} 73.9 & {\cellcolor[HTML]{E8F6E5}} 85.4 \\
Mistral Large 3 & {\cellcolor[HTML]{98CB9D}} 86.7 & {\cellcolor[HTML]{9ECFA2}} 71.2 & {\cellcolor[HTML]{6C9B7D}} 77.8 & {\cellcolor[HTML]{C0E4BE}} 83.1 & {\cellcolor[HTML]{EAF6E6}} 46.4 & {\cellcolor[HTML]{9CCEA0}} 84.5 & {\cellcolor[HTML]{72A683}} 83.5 & {\cellcolor[HTML]{B3DCB3}} 83.5 & {\cellcolor[HTML]{92C699}} 85.3 & {\cellcolor[HTML]{70A180}} 85.6 & {\cellcolor[HTML]{99CC9E}} 85.9 & {\cellcolor[HTML]{668F76}} 80.3 & {\cellcolor[HTML]{BCE1BB}} 87.6 \\
Qwen 3.5 397B & {\cellcolor[HTML]{CEEBCA}} 83.2 & {\cellcolor[HTML]{699579}} 79.2 & {\cellcolor[HTML]{6C9A7C}} 78.3 & {\cellcolor[HTML]{EEF7EB}} 79.7 & {\cellcolor[HTML]{668F76}} 70.9 & {\cellcolor[HTML]{CEEBCA}} 81.2 & {\cellcolor[HTML]{6E9F7F}} 84.7 & {\cellcolor[HTML]{AAD7AC}} 83.9 & {\cellcolor[HTML]{BEE3BC}} 83.1 & {\cellcolor[HTML]{73A884}} 84.8 & {\cellcolor[HTML]{8FC397}} 86.5 & {\cellcolor[HTML]{7DB48B}} 75.9 & {\cellcolor[HTML]{D9F0D5}} 86.2 \\
DeepL NextGen & {\cellcolor[HTML]{95C89B}} 87.0 & {\cellcolor[HTML]{668F76}} 80.1 & {\cellcolor[HTML]{668F76}} 80.8 & {\cellcolor[HTML]{F8F6F8}} 77.7 & {\cellcolor[HTML]{6A967A}} 69.4 & {\cellcolor[HTML]{73A884}} 87.8 & {\cellcolor[HTML]{8FC397}} 77.8 & {\cellcolor[HTML]{E7D9EB}} 74.3 & {\cellcolor[HTML]{D0ECCD}} 82.1 & {\cellcolor[HTML]{6FA07F}} 85.8 & {\cellcolor[HTML]{BEE3BC}} 84.3 & {\cellcolor[HTML]{95C89B}} 73.5 & {\cellcolor[HTML]{E7F6E3}} 85.5 \\
Gemma 4 31b & {\cellcolor[HTML]{668F76}} 92.1 & {\cellcolor[HTML]{D3C0DC}} 48.5 & {\cellcolor[HTML]{D4EED0}} 56.8 & {\cellcolor[HTML]{668F76}} 89.2 & {\cellcolor[HTML]{E2D2E7}} 30.6 & {\cellcolor[HTML]{71A582}} 88.1 & {\cellcolor[HTML]{668F76}} 88.1 & {\cellcolor[HTML]{668F76}} 89.4 & {\cellcolor[HTML]{668F76}} 88.9 & {\cellcolor[HTML]{7DB48B}} 83.3 & {\cellcolor[HTML]{668F76}} 89.9 & {\cellcolor[HTML]{C2E5BF}} 69.8 & {\cellcolor[HTML]{78B088}} 90.4 \\
Inkling Small & {\cellcolor[HTML]{C2E5BF}} 84.1 & {\cellcolor[HTML]{B0DAB1}} 69.6 & {\cellcolor[HTML]{75AC86}} 73.9 & {\cellcolor[HTML]{F1F8F0}} 79.3 & {\cellcolor[HTML]{6D9C7D}} 68.2 & {\cellcolor[HTML]{C7E8C4}} 81.7 & {\cellcolor[HTML]{8ABF94}} 78.5 & {\cellcolor[HTML]{D3EDCF}} 81.6 & {\cellcolor[HTML]{E5F5E1}} 80.7 & {\cellcolor[HTML]{9CCEA0}} 80.1 & {\cellcolor[HTML]{F7F9F6}} 80.1 & {\cellcolor[HTML]{F5F9F4}} 63.0 & {\cellcolor[HTML]{DBF1D7}} 86.1 \\
Muse Glimmer 30B & {\cellcolor[HTML]{D4EED0}} 82.7 & {\cellcolor[HTML]{BCE1BB}} 68.4 & {\cellcolor[HTML]{74AA85}} 74.4 & {\cellcolor[HTML]{E3F4E0}} 80.7 & {\cellcolor[HTML]{A1D2A5}} 58.0 & {\cellcolor[HTML]{D1EDCE}} 80.8 & {\cellcolor[HTML]{7EB58C}} 80.5 & {\cellcolor[HTML]{E3F4E0}} 80.4 & {\cellcolor[HTML]{F9FAF9}} 78.6 & {\cellcolor[HTML]{92C699}} 81.1 & {\cellcolor[HTML]{F0F8EE}} 80.9 & {\cellcolor[HTML]{CEEBCA}} 68.7 & {\cellcolor[HTML]{E8F6E5}} 85.4 \\
\cellcolor[HTML]{D9F0D3}Command A+ & {\cellcolor[HTML]{E5F5E1}} 81.1 & {\cellcolor[HTML]{D3EDCF}} 65.7 & {\cellcolor[HTML]{A0D1A3}} 65.7 & {\cellcolor[HTML]{FAFAFA}} 78.2 & {\cellcolor[HTML]{CAEAC7}} 52.6 & {\cellcolor[HTML]{92C699}} 85.2 & {\cellcolor[HTML]{89BE93}} 78.7 & {\cellcolor[HTML]{ECF7E9}} 79.5 & {\cellcolor[HTML]{ECF7EA}} 80.0 & {\cellcolor[HTML]{81B88E}} 82.9 & {\cellcolor[HTML]{F9F8F9}} 79.5 & {\cellcolor[HTML]{8ABF94}} 74.6 & {\cellcolor[HTML]{E7D9EB}} 81.3 \\
GLM 5.2 & {\cellcolor[HTML]{A3D3A6}} 86.0 & {\cellcolor[HTML]{86BB91}} 73.8 & {\cellcolor[HTML]{6C9A7C}} 78.0 & {\cellcolor[HTML]{F1F8F0}} 79.3 & {\cellcolor[HTML]{9CCEA0}} 58.6 & {\cellcolor[HTML]{A5D4A8}} 83.9 & {\cellcolor[HTML]{BC9EC6}} 37.0 & {\cellcolor[HTML]{8CC095}} 85.7 & {\cellcolor[HTML]{D0BCD9}} 73.4 & {\cellcolor[HTML]{7DB48B}} 83.3 & {\cellcolor[HTML]{72A683}} 88.4 & {\cellcolor[HTML]{84BA90}} 75.2 & {\cellcolor[HTML]{668F76}} 92.1 \\
Mistral Medium 3.5 & {\cellcolor[HTML]{E0F3DC}} 81.5 & {\cellcolor[HTML]{F2E9F3}} 55.0 & {\cellcolor[HTML]{C9E9C5}} 59.3 & {\cellcolor[HTML]{F5EFF5}} 77.0 & {\cellcolor[HTML]{F1E5F1}} 34.7 & {\cellcolor[HTML]{C0E4BE}} 82.2 & {\cellcolor[HTML]{95C89B}} 77.0 & {\cellcolor[HTML]{DCF1D9}} 81.0 & {\cellcolor[HTML]{E0F3DC}} 81.0 & {\cellcolor[HTML]{98CB9D}} 80.5 & {\cellcolor[HTML]{DAF0D6}} 82.8 & {\cellcolor[HTML]{87BD92}} 74.9 & {\cellcolor[HTML]{CFECCB}} 86.8 \\
GPT-OSS 120B & {\cellcolor[HTML]{F5F9F4}} 78.8 & {\cellcolor[HTML]{F6F9F6}} 59.4 & {\cellcolor[HTML]{B1DBB2}} 62.9 & {\cellcolor[HTML]{F4ECF4}} 76.7 & {\cellcolor[HTML]{E2F4DF}} 48.0 & {\cellcolor[HTML]{EDF7EB}} 77.9 & {\cellcolor[HTML]{A7D5A9}} 74.2 & {\cellcolor[HTML]{C8B1D2}} 71.3 & {\cellcolor[HTML]{E2D2E7}} 74.9 & {\cellcolor[HTML]{C3E6C1}} 76.6 & {\cellcolor[HTML]{CCB7D6}} 74.6 & {\cellcolor[HTML]{F9F6F9}} 61.2 & {\cellcolor[HTML]{E8DAEB}} 81.5 \\
\cellcolor[HTML]{D9F0D3}Command A Translate (2025) & {\cellcolor[HTML]{F1F8EF}} 79.5 & {\cellcolor[HTML]{DCCBE3}} 49.9 & {\cellcolor[HTML]{C5E7C2}} 59.7 & {\cellcolor[HTML]{C6AED0}} 71.9 & {\cellcolor[HTML]{F3EBF4}} 36.3 & {\cellcolor[HTML]{F5F0F6}} 74.3 & {\cellcolor[HTML]{E8DAEB}} 48.7 & {\cellcolor[HTML]{EAF6E6}} 79.8 & {\cellcolor[HTML]{ECF7EA}} 80.0 & {\cellcolor[HTML]{83B98F}} 82.7 & {\cellcolor[HTML]{F4EDF4}} 78.4 & {\cellcolor[HTML]{D0BCD9}} 53.0 & {\cellcolor[HTML]{E1D1E7}} 80.9 \\
Google Translate & {\cellcolor[HTML]{DDCCE4}} 72.6 & {\cellcolor[HTML]{D7EFD4}} 65.0 & {\cellcolor[HTML]{B3DCB3}} 62.7 & {\cellcolor[HTML]{8C6693}} 67.1 & {\cellcolor[HTML]{6C9B7D}} 68.3 & {\cellcolor[HTML]{E3D4E8}} 71.1 & {\cellcolor[HTML]{DFF2DB}} 65.2 & {\cellcolor[HTML]{8C6693}} 66.0 & {\cellcolor[HTML]{936B9A}} 68.4 & {\cellcolor[HTML]{F8F6F8}} 67.3 & {\cellcolor[HTML]{8C6693}} 69.6 & {\cellcolor[HTML]{E0F3DC}} 66.6 & {\cellcolor[HTML]{8C6693}} 75.4 \\
Nemotron 3 Ultra 550B & {\cellcolor[HTML]{8FC397}} 87.4 & {\cellcolor[HTML]{C9B2D3}} 46.7 & {\cellcolor[HTML]{EEF7EB}} 49.6 & {\cellcolor[HTML]{AF83B7}} 69.6 & {\cellcolor[HTML]{E5D7EA}} 31.6 & {\cellcolor[HTML]{8C6693}} 61.3 & {\cellcolor[HTML]{C7E8C4}} 69.9 & {\cellcolor[HTML]{916A98}} 66.4 & {\cellcolor[HTML]{8C6693}} 67.9 & {\cellcolor[HTML]{BB9CC5}} 54.9 & {\cellcolor[HTML]{EEE2EF}} 77.5 & {\cellcolor[HTML]{8C6693}} 43.5 & {\cellcolor[HTML]{C2E5BF}} 87.4 \\
Qwen 3.6 27B & {\cellcolor[HTML]{8C6693}} 63.9 & {\cellcolor[HTML]{8C6693}} 36.7 & {\cellcolor[HTML]{8C6693}} 8.7 & {\cellcolor[HTML]{ECF7EA}} 79.8 & {\cellcolor[HTML]{8C6693}} 10.6 & {\cellcolor[HTML]{F8F5F8}} 75.0 & {\cellcolor[HTML]{8C6693}} 26.2 & {\cellcolor[HTML]{CDEBC9}} 82.1 & {\cellcolor[HTML]{F7F4F8}} 77.8 & {\cellcolor[HTML]{8C6693}} 48.1 & {\cellcolor[HTML]{A9D6AB}} 85.3 & {\cellcolor[HTML]{F4EDF4}} 59.6 & {\cellcolor[HTML]{71A481}} 91.0 \\
\bottomrule
\end{tabular}

\begin{tabular}{lrrrrrrrrrrrrr}
\toprule
 & \makecell{Slovak} & \makecell{Slovenian} & \makecell{Spanish} & \makecell{Standard\\Arabic} & \makecell{Swedish} & \makecell{Tamil} & \makecell{Telugu} & \makecell{Thai} & \makecell{Traditional\\Chinese} & \makecell{Turkish} & \makecell{Ukrainian} & \makecell{Urdu} & \makecell{Vietnamese} \\
\midrule
\cellcolor[HTML]{D9F0D3}North Small Translate (Agentic) & {\cellcolor[HTML]{668F76}} 84.6 & {\cellcolor[HTML]{87BD92}} 76.1 & {\cellcolor[HTML]{668F76}} 90.9 & {\cellcolor[HTML]{6FA07F}} 86.6 & {\cellcolor[HTML]{668F76}} 89.2 & {\cellcolor[HTML]{668F76}} 85.7 & {\cellcolor[HTML]{6A977B}} 85.9 & {\cellcolor[HTML]{77AF87}} 86.7 & {\cellcolor[HTML]{699579}} 88.8 & {\cellcolor[HTML]{70A180}} 85.3 & {\cellcolor[HTML]{6C9B7D}} 86.8 & {\cellcolor[HTML]{6C9A7C}} 87.1 & {\cellcolor[HTML]{6C9B7D}} 89.7 \\
\cellcolor[HTML]{D9F0D3}North Small Translate & {\cellcolor[HTML]{86BB91}} 81.2 & {\cellcolor[HTML]{7DB48B}} 77.2 & {\cellcolor[HTML]{6A977B}} 90.5 & {\cellcolor[HTML]{6E9D7E}} 86.7 & {\cellcolor[HTML]{689278}} 88.9 & {\cellcolor[HTML]{668F76}} 85.7 & {\cellcolor[HTML]{6B997B}} 85.7 & {\cellcolor[HTML]{77AF87}} 86.7 & {\cellcolor[HTML]{6A977B}} 88.5 & {\cellcolor[HTML]{6E9D7E}} 85.7 & {\cellcolor[HTML]{6B997B}} 86.9 & {\cellcolor[HTML]{6D9C7D}} 86.9 & {\cellcolor[HTML]{70A180}} 89.4 \\
Mistral Large 3 & {\cellcolor[HTML]{A5D4A8}} 79.3 & {\cellcolor[HTML]{98CB9D}} 74.5 & {\cellcolor[HTML]{92C699}} 88.4 & {\cellcolor[HTML]{C9E9C5}} 80.2 & {\cellcolor[HTML]{D1EDCE}} 82.7 & {\cellcolor[HTML]{71A582}} 82.2 & {\cellcolor[HTML]{75AB85}} 82.5 & {\cellcolor[HTML]{90C498}} 84.2 & {\cellcolor[HTML]{6D9C7D}} 87.8 & {\cellcolor[HTML]{78B088}} 82.9 & {\cellcolor[HTML]{83B98F}} 84.7 & {\cellcolor[HTML]{A7D5A9}} 80.8 & {\cellcolor[HTML]{A7D5A9}} 86.6 \\
Qwen 3.5 397B & {\cellcolor[HTML]{AAD7AC}} 79.1 & {\cellcolor[HTML]{699579}} 81.0 & {\cellcolor[HTML]{C5E7C2}} 86.7 & {\cellcolor[HTML]{77AF87}} 85.4 & {\cellcolor[HTML]{AAD7AC}} 84.6 & {\cellcolor[HTML]{6C9A7C}} 83.9 & {\cellcolor[HTML]{71A481}} 83.8 & {\cellcolor[HTML]{AAD7AC}} 82.0 & {\cellcolor[HTML]{87BD92}} 82.8 & {\cellcolor[HTML]{73A884}} 84.0 & {\cellcolor[HTML]{73A783}} 85.9 & {\cellcolor[HTML]{C9E9C5}} 78.2 & {\cellcolor[HTML]{A7D5A9}} 86.5 \\
DeepL NextGen & {\cellcolor[HTML]{6A967A}} 84.0 & {\cellcolor[HTML]{668F76}} 81.9 & {\cellcolor[HTML]{92C699}} 88.4 & {\cellcolor[HTML]{AAD7AC}} 82.0 & {\cellcolor[HTML]{6C9A7C}} 88.4 & {\cellcolor[HTML]{8ABF94}} 77.9 & {\cellcolor[HTML]{84BA90}} 79.8 & {\cellcolor[HTML]{AAD7AC}} 81.9 & {\cellcolor[HTML]{7AB189}} 84.6 & {\cellcolor[HTML]{70A180}} 85.1 & {\cellcolor[HTML]{8DC296}} 84.1 & {\cellcolor[HTML]{DFF2DB}} 75.7 & {\cellcolor[HTML]{A3D3A6}} 86.7 \\
Gemma 4 31b & {\cellcolor[HTML]{F4EDF4}} 71.2 & {\cellcolor[HTML]{F6F9F6}} 63.1 & {\cellcolor[HTML]{6A967A}} 90.6 & {\cellcolor[HTML]{668F76}} 88.1 & {\cellcolor[HTML]{699479}} 88.8 & {\cellcolor[HTML]{668F76}} 85.8 & {\cellcolor[HTML]{668F76}} 87.5 & {\cellcolor[HTML]{668F76}} 90.5 & {\cellcolor[HTML]{6E9F7F}} 87.4 & {\cellcolor[HTML]{668F76}} 88.1 & {\cellcolor[HTML]{668F76}} 87.7 & {\cellcolor[HTML]{668F76}} 88.5 & {\cellcolor[HTML]{668F76}} 90.4 \\
Inkling Small & {\cellcolor[HTML]{C2E5BF}} 77.8 & {\cellcolor[HTML]{83B98F}} 76.6 & {\cellcolor[HTML]{E7F6E3}} 85.2 & {\cellcolor[HTML]{D9F0D5}} 78.9 & {\cellcolor[HTML]{B7DEB6}} 84.0 & {\cellcolor[HTML]{9CCEA0}} 75.6 & {\cellcolor[HTML]{93C79A}} 77.7 & {\cellcolor[HTML]{D4EED0}} 78.1 & {\cellcolor[HTML]{92C699}} 81.6 & {\cellcolor[HTML]{A7D5A9}} 77.2 & {\cellcolor[HTML]{B5DDB5}} 82.0 & {\cellcolor[HTML]{DAF0D6}} 76.3 & {\cellcolor[HTML]{EFF8ED}} 82.8 \\
Muse Glimmer 30B & {\cellcolor[HTML]{C7E8C4}} 77.5 & {\cellcolor[HTML]{9CCEA0}} 74.0 & {\cellcolor[HTML]{EEF7EB}} 84.7 & {\cellcolor[HTML]{F8F5F8}} 74.3 & {\cellcolor[HTML]{EBF7E8}} 80.9 & {\cellcolor[HTML]{7DB48B}} 79.7 & {\cellcolor[HTML]{8CC095}} 78.7 & {\cellcolor[HTML]{E6F5E2}} 76.1 & {\cellcolor[HTML]{8ABF94}} 82.5 & {\cellcolor[HTML]{9ECFA2}} 78.2 & {\cellcolor[HTML]{CAEAC7}} 80.9 & {\cellcolor[HTML]{B9DFB8}} 79.3 & {\cellcolor[HTML]{F0F8EE}} 82.8 \\
\cellcolor[HTML]{D9F0D3}Command A+ & {\cellcolor[HTML]{CDEBC9}} 77.2 & {\cellcolor[HTML]{9CCEA0}} 73.9 & {\cellcolor[HTML]{B7DEB6}} 87.2 & {\cellcolor[HTML]{AAD7AC}} 82.0 & {\cellcolor[HTML]{CDEBC9}} 83.0 & {\cellcolor[HTML]{95C89B}} 76.5 & {\cellcolor[HTML]{A0D1A3}} 75.9 & {\cellcolor[HTML]{D4EED0}} 78.3 & {\cellcolor[HTML]{CFBBD8}} 53.6 & {\cellcolor[HTML]{A1D2A5}} 77.8 & {\cellcolor[HTML]{ECF7E9}} 78.4 & {\cellcolor[HTML]{D0ECCD}} 77.4 & {\cellcolor[HTML]{F9FAF9}} 81.9 \\
GLM 5.2 & {\cellcolor[HTML]{6D9C7D}} 83.6 & {\cellcolor[HTML]{A0D1A3}} 73.8 & {\cellcolor[HTML]{E8DAEB}} 81.7 & {\cellcolor[HTML]{6E9F7F}} 86.7 & {\cellcolor[HTML]{E7F6E3}} 81.3 & {\cellcolor[HTML]{CAB3D3}} 46.7 & {\cellcolor[HTML]{8C6693}} 31.8 & {\cellcolor[HTML]{8CC095}} 84.6 & {\cellcolor[HTML]{6A967A}} 88.8 & {\cellcolor[HTML]{6FA07F}} 85.3 & {\cellcolor[HTML]{92C699}} 83.8 & {\cellcolor[HTML]{80B68D}} 84.1 & {\cellcolor[HTML]{8CC095}} 87.7 \\
Mistral Medium 3.5 & {\cellcolor[HTML]{F5EEF5}} 71.3 & {\cellcolor[HTML]{EBF7E8}} 65.6 & {\cellcolor[HTML]{A1D2A5}} 87.9 & {\cellcolor[HTML]{E6F5E2}} 77.8 & {\cellcolor[HTML]{CFECCB}} 82.9 & {\cellcolor[HTML]{AAD7AC}} 73.9 & {\cellcolor[HTML]{B0DAB1}} 73.9 & {\cellcolor[HTML]{CDEBC9}} 79.0 & {\cellcolor[HTML]{7DB48B}} 84.2 & {\cellcolor[HTML]{A1D2A5}} 77.8 & {\cellcolor[HTML]{BAE0B9}} 81.8 & {\cellcolor[HTML]{E2F4DF}} 75.3 & {\cellcolor[HTML]{E3F4E0}} 83.8 \\
GPT-OSS 120B & {\cellcolor[HTML]{E5D7EA}} 69.0 & {\cellcolor[HTML]{E5F5E1}} 66.6 & {\cellcolor[HTML]{F1E6F2}} 82.3 & {\cellcolor[HTML]{F2E9F3}} 72.9 & {\cellcolor[HTML]{F2E9F3}} 77.7 & {\cellcolor[HTML]{CBEAC8}} 70.1 & {\cellcolor[HTML]{BAE0B9}} 72.7 & {\cellcolor[HTML]{FAFAFA}} 72.1 & {\cellcolor[HTML]{8FC397}} 81.9 & {\cellcolor[HTML]{CAEAC7}} 73.5 & {\cellcolor[HTML]{DDCDE4}} 72.5 & {\cellcolor[HTML]{E2F4DF}} 75.3 & {\cellcolor[HTML]{A579AC}} 74.6 \\
\cellcolor[HTML]{D9F0D3}Command A Translate (2025) & {\cellcolor[HTML]{F6F9F5}} 73.3 & {\cellcolor[HTML]{F3EBF4}} 59.4 & {\cellcolor[HTML]{DFF2DB}} 85.6 & {\cellcolor[HTML]{AAD7AC}} 82.0 & {\cellcolor[HTML]{D2BEDB}} 74.7 & {\cellcolor[HTML]{E2F4DF}} 66.4 & {\cellcolor[HTML]{C8B1D2}} 44.2 & {\cellcolor[HTML]{C3AACD}} 61.2 & {\cellcolor[HTML]{A7D5A9}} 79.1 & {\cellcolor[HTML]{B0DAB1}} 76.3 & {\cellcolor[HTML]{E2F4DF}} 79.3 & {\cellcolor[HTML]{A579AC}} 56.6 & {\cellcolor[HTML]{EEE2EF}} 79.9 \\
Google Translate & {\cellcolor[HTML]{C3AACD}} 65.6 & {\cellcolor[HTML]{ECF7E9}} 65.4 & {\cellcolor[HTML]{8C6693}} 76.5 & {\cellcolor[HTML]{F3EAF3}} 73.0 & {\cellcolor[HTML]{8C6693}} 69.4 & {\cellcolor[HTML]{C2E5BF}} 71.4 & {\cellcolor[HTML]{8FC397}} 78.3 & {\cellcolor[HTML]{E2F4DF}} 76.5 & {\cellcolor[HTML]{8C6693}} 41.3 & {\cellcolor[HTML]{DDF2DA}} 70.4 & {\cellcolor[HTML]{8C6693}} 65.5 & {\cellcolor[HTML]{8C6693}} 54.0 & {\cellcolor[HTML]{DFCFE5}} 78.8 \\
Nemotron 3 Ultra 550B & {\cellcolor[HTML]{8C6693}} 60.7 & {\cellcolor[HTML]{E5D7EA}} 56.3 & {\cellcolor[HTML]{99CC9E}} 88.1 & {\cellcolor[HTML]{CEB9D8}} 68.5 & {\cellcolor[HTML]{B28ABB}} 71.9 & {\cellcolor[HTML]{ECE0EE}} 54.1 & {\cellcolor[HTML]{E9DBEC}} 52.1 & {\cellcolor[HTML]{DCCBE3}} 65.0 & {\cellcolor[HTML]{80B68D}} 83.8 & {\cellcolor[HTML]{ECF7EA}} 67.6 & {\cellcolor[HTML]{8F6896}} 65.7 & {\cellcolor[HTML]{F9F6F9}} 70.6 & {\cellcolor[HTML]{8C6693}} 73.3 \\
Qwen 3.6 27B & {\cellcolor[HTML]{F3EAF3}} 70.9 & {\cellcolor[HTML]{8C6693}} 42.5 & {\cellcolor[HTML]{E6D8EA}} 81.6 & {\cellcolor[HTML]{8C6693}} 61.9 & {\cellcolor[HTML]{D1BDDA}} 74.5 & {\cellcolor[HTML]{8C6693}} 34.7 & {\cellcolor[HTML]{B7DEB6}} 73.2 & {\cellcolor[HTML]{8C6693}} 53.7 & {\cellcolor[HTML]{668F76}} 89.9 & {\cellcolor[HTML]{8C6693}} 39.7 & {\cellcolor[HTML]{BFA4CA}} 69.7 & {\cellcolor[HTML]{DDF2DA}} 75.9 & {\cellcolor[HTML]{D5EED1}} 84.6 \\
\bottomrule
\end{tabular}
\end{minipage}%
}
\caption{Machine translation quality of every supported language of the WMT26 testset.}
\label{app:per_language}
\vspace{-1em}
\end{table*}

\clearpage

\section{GEMBA-ESA Fluency+Adequacy Template}
\label{app:gemba-esa-fluade}

\subsection{System prompt}

\begin{Verbatim}[fontsize=\scriptsize,breaklines=true,frame=single]
You are an impartial judge whose task is to analyze machine translation outputs and evaluate their quality.

Based on the provided source text, target language, translation guidelines, and translation output, identify translation errors and assess their severity. Then evaluate the quality of the translation according to the provided rubric criteria. For each criterion, provide a brief rationale in English and a score within the allowed range.

Errors have "quote", "severity", and "category" attributes. The "quote" value should quote the part of the translation that contains the error. Use minor severity for imperfections or stylistic issues that do not impact the core message (e.g., awkward phrasing). Use major severity for errors that obscure the meaning, misrepresent the source, or change the message (e.g., incorrect information or confusing wording). The category must be one of: "accuracy", "fluency", "terminology", "style", "locale", "design", "mistranslation", "omission", "addition", "untranslated", "grammar", "spelling", "punctuation", "register", "unidiomatic", "inconsistency", or "other". If the error span is longer than a few words, use an ellipsis (...) to shorten it. If the error concerns the entire translation, such as a translation in the wrong language, set "quote" to "FULL_TRANSLATION". For omissions (missing source content), set "quote" to the adjacent translation words with "OMISSION" between them. If there are no errors, set "errors" to an empty array.

Return your error analysis and quality evaluation as a single, strictly valid JSON object matching the provided output format.
\end{Verbatim}

\subsection{User prompt}

\begin{Verbatim}[fontsize=\scriptsize,breaklines=true,frame=single]
## Evaluation rubric
- Fluency score [key: fluency]: Measures how natural and grammatically correct the translation reads in the target language. (score: 0 to 100)
- Adequacy score [key: adequacy]: Measures how completely the meaning of the source is conveyed by the translation. (score: 0 to 100)

## Additional judging rules
### Important rules
- Hallucinations are major adequacy errors and should be penalized severely.
- Outputs in the wrong language, dialect, or script are major fluency errors and should be penalized severely. Borrowed words are allowed if valid in the target language.
- Check translation consistency (e.g., technical terms) across the entire text.
- Over-translation and extraneous text (e.g., introductory statements like "Here is your translation" or unprompted explanations) are major adequacy errors.
- Count guideline adherence issues toward the adequacy score; for these errors, set "quote" to the corresponding guideline.

### Adequacy scale
- 81-100 (Very Good): Complete meaning transfer; no information is lost, added, or distorted.
- 61-80 (Good): Near-complete transfer; minor inaccuracies or omissions that do not affect the core message.
- 41-60 (Acceptable): Main ideas conveyed, but noticeable inaccuracies, omissions, or additions.
- 21-40 (Borderline): Partial transfer; frequent misinterpretation or omission confusing the core message.
- 0-20 (Not acceptable): Complete violation of meaning; large portions mistranslated, hallucinated, or missing.

### Fluency scale
- 81-100 (Very Good): Perfectly natural and grammatical; reads like a native text; requires no or minimal proofreading.
- 61-80 (Good): Mostly natural; minor awkwardness or slight grammatical imperfections; needs light proofreading.
- 41-60 (Acceptable): Uneven naturalness; noticeable awkward phrasing or structural issues; usable only after substantial revision.
- 21-40 (Borderline): Often unnatural; frequent grammatical errors that impede reading flow; requires major rewrite.
- 0-20 (Not acceptable): Incoherent, structurally broken, incomprehensible, or in the wrong language, dialect, or script; unusable without complete retranslation.

## Source text
```
{{SOURCE_TEXT}}
```

## Translation language or dialect
{{TARGET_LANGUAGE_OR_DIALECT}}

## Translation guidelines
{{TRANSLATION_GUIDELINES}}

## Translation output
```
{{TRANSLATION_OUTPUT}}
```

## Expected output format
Return a single, strictly valid JSON object with the following schema:
```json
{
  "errors": [
    {
      "quote": "<string>",
      "severity": "minor|major",
      "category": "<string>"
    }
  ],
  "fluency": {
    "rationale": "<string>",
    "score": "<number>"
  },
  "adequacy": {
    "rationale": "<string>",
    "score": "<number>"
  }
}
```

Every string value must be valid JSON: escape each backslash as \\ and each double quote as \". Do not emit raw control characters or trailing commas.
\end{Verbatim}

\end{document}